\documentclass[10pt,journal,compsoc]{IEEEtran}

\ifCLASSOPTIONcompsoc
  \usepackage[nocompress]{cite}
\else
  \usepackage{cite}
\fi

\usepackage{color}
\usepackage{xcolor}
\usepackage{graphicx}
\usepackage{amsmath}
\usepackage{amssymb}
\usepackage{epsfig}
\usepackage{algorithmic}

\usepackage{soul}
\usepackage[]{footmisc}

\newcommand{\cell}[1]{\begin{tabular}{l} #1 \end{tabular}}

\begin{document}
%



\title{Dual networks based 3D Multi-Person Pose Estimation from Monocular Video}

%
%
%
%

\author{Yu~Cheng,
        Bo~Wang,
        and~Robby~T.~Tan,~\IEEEmembership{Member,~IEEE}%
        
\IEEEcompsocitemizethanks{\IEEEcompsocthanksitem Yu Cheng and Robby T. Tan are with Yale-NUS College and the Department of Electrical
and Computer Engineering, National University of Singapore, Singapore. E-mail: e0321276@u.nus.edu, robby.tan@nus.edu.sg

\IEEEcompsocthanksitem Bo Wang is with CtrsVision (USA). E-mail: hawk.rsrch@gmail.com}

\thanks{Manuscript received May 31, 2021.}}

\markboth{IEEE TRANSACTIONS ON PATTERN ANALYSIS AND MACHINE INTELLIGENCE, VOL. XXX, NO. XXX, May YYYY}%
{Shell \MakeLowercase{\textit{et al.}}: Bare Demo of IEEEtran.cls for Computer Society Journals}
%



\IEEEtitleabstractindextext{%
\begin{abstract}
Monocular 3D human pose estimation has made progress in  recent years. Most of the methods focus on single persons, which estimate the poses in the person-centric coordinates, i.e., the coordinates based on the center of the target person. Hence, these methods are inapplicable for multi-person 3D pose estimation, where the absolute coordinates (e.g., the camera coordinates) are required. Moreover, multi-person pose estimation is more challenging than single pose estimation, due to inter-person occlusion and close human interactions.
Existing top-down multi-person methods rely on human detection (i.e., top-down approach), and thus suffer from the detection errors and cannot produce reliable pose estimation in multi-person scenes. 
Meanwhile, existing bottom-up methods that do not use human detection are not affected by detection errors, but since they process all persons in a scene at once, they are prone to errors, particularly for persons in small scales.
To address all these challenges, we propose the integration of top-down and bottom-up approaches to exploit their strengths. Our top-down network estimates human joints from all persons instead of one in an image patch, making it robust to possible erroneous bounding boxes. Our bottom-up network incorporates human-detection based normalized heatmaps, allowing the network to be more robust in handling scale variations. Finally, the estimated 3D poses from the top-down and bottom-up networks are fed into our integration network for final 3D poses. 
To address the common gaps between training and testing data, we do optimization during the test time, by refining the estimated 3D human poses using high-order temporal constraint, re-projection loss, and bone length regularizations.
We also introduce a two-person pose discriminator that enforces natural two-person interactions. Finally, we apply a semi-supervised method to overcome the 3D ground-truth data scarcity.
Our evaluations demonstrate the effectiveness of the proposed method and its individual components. Our code and pretrained models are available publicly: https://github.com/3dpose/3D-Multi-Person-Pose.
\end{abstract}

}

\maketitle

\IEEEdisplaynontitleabstractindextext

%
\IEEEpeerreviewmaketitle

\IEEEraisesectionheading{\section{Introduction}\label{sec:introduction}}

\begin{table*}
\footnotesize
\centering
  \begin{tabular}{@{}c@{} @{}c@{} @{}l@{} @{}c@{}}
  \cline{1-4}
  \cline{1-4}
    \rule{0pt}{3.5ex}
    \cell{\textbf{Top-down /} \\ \textbf{Bottom-up}} & \textbf{2D / 3D} & \textbf{Coordinate system} & \textbf{Methods} \\
    \cline{1-4}
    \rule{0pt}{4ex}
    \cell{Top-down} & \cell{2D human pose \\ estimation} & 2D image coordinate & \cell{\cite{he2017mask,papandreou2017towards,liu2018cascaded,sun2018integral,chen2018cascaded,fang2017rmpe,huang2017coarse,xiao2018simple,sun2019hrnet,li2019crowdpose,wang2020deep}}\\
    \cline{2-4}
    \rule{0pt}{3.5ex}
    & \cell{3D human pose \\ estimation} & 3D person-centric & \cell{\cite{zhou2016sparseness,martinez2017simple,mehta2017monocular,zhou2017towards,mehta2017vnect,chen20173d,li2019generating,nie2017monocular,li2020cascaded,ci2020locally,Yang20183DHP,pavlakos2018ordinal} \\ \cite{hossain2018exploiting,pavllo20193d,wandt2019repnet,nibali20193d,cheng2019occlusion,sharma2019monocular,arnab2019exploiting,cheng2020sptaiotemporal}}\\
    \cline{3-4}
    \rule{0pt}{3.5ex}
    & & 3D camera-centric & \cell{\cite{rogez2019lcr,bertoni2019monoloco,mehta2020xnect,Moon_2019_ICCV_3DMPPE,li2020hmor,lin2020hdnet,cheng2021graph}}\\
    \cline{1-4}
    \rule{0pt}{3.5ex}
    \cell{Bottom-up} & \cell{2D human pose \\ estimation} & 2D image coordinate & \cell{\cite{pishchulin2016deepcut,insafutdinov2016deepercut,cao2017realtime,newell2017associative,insafutdinov2017arttrack,iqbal2017posetrack,papandreou2018personlab,cao2019openpose,jin2019multi,kreiss2019pifpaf,cheng2020higherhrnet,jin2020differentiable} \\
    \cite{geng2021bottom,Luo_2021_CVPR}}\\
    \cline{2-4}
    \rule{0pt}{3.5ex}
    & \cell{3D human pose \\ estimation} & 3D person-centric & \cell{N/A}\\
    \cline{3-4}
    \rule{0pt}{3.5ex}
    & & 3D camera-centric & \cell{\cite{benzine2020pandanet,fabbri2020compressed,zhen2020smap}}\\
    \cline{1-4}
  \end{tabular}
  \vspace{0.5em}
  \caption{{Summary of the top-down and bottom-up 2D/3D human pose estimation methods and the coordinate systems of the obtained human pose results.}}
  \label{tab:summary_td_bu}
\end{table*}

\begin{table*}
\footnotesize
\centering
  \begin{tabular}{@{}c@{} @{}l@{} @{}l@{} @{}l@{} @{}l@{}}
  \cline{1-5}
  \cline{1-5}
    \rule{0pt}{2.6ex}
    \textbf{Task} & \textbf{Coordinate system} & \textbf{Metric} & \textbf{Dataset} & \textbf{Publication} \\
    \cline{1-5}
    \rule{0pt}{6ex}
    \cell{3D human pose} & \cell{Person-centric \\ (relative coordinate)} & \cell{MPJPE, \\ PA-MPJPE,\\ PCK, \\ $AUC_{rel}$} & \cell{Human3.6M, \\ 3DHP,\\ HumanEva, \\ Penn Action} & \cell{ \cite{zhou2016sparseness,martinez2017simple,mehta2017monocular,zhou2017towards,mehta2017vnect} \cite{chen20173d, li2019generating, nie2017monocular,li2020cascaded,ci2020locally} \\
    \cite{Yang20183DHP,pavlakos2018ordinal,hossain2018exploiting,pavllo20193d,wandt2019repnet} \cite{nibali20193d,cheng2019occlusion,sharma2019monocular,arnab2019exploiting,cheng2020sptaiotemporal} }\\
    \cline{1-5}
    \rule{0pt}{6ex}
    \cell{3D multi-person pose} & \cell{Camera-centric \\ (absolute coordinate)} & \cell{PCK$_{abs}$,\\ $AP_{25}^{root}$,\\ MPRE~\cite{Moon_2019_ICCV_3DMPPE}, \\ F1 value~\cite{fabbri2020compressed}} & \cell{MuPoTS-3D,\\ JTA,\\ 3DPW} & \cell{\cite{mehta2018single,rogez2019lcr,Moon_2019_ICCV_3DMPPE,nie2019single,bertoni2019monoloco,mehta2020xnect,fabbri2020compressed,benzine2020pandanet,li2020hmor,lin2020hdnet} \\ \cite{zhen2020smap,cheng2021graph}}\\
    \cline{1-5}
  \end{tabular}
  \vspace{0.5em}
  \caption{Summary of the differences between 3D human pose estimation and multi-person pose estimation.}
  \label{tab:summary}
\end{table*}

3D multi-person pose estimation from a monocular video is useful for many real-world applications (e.g., \cite{desmarais2020review,luvizon2020multi,wang2020vr,srivastav2018mvor,tuyls2020game}).
However, this multi-person estimation is challenging not only because of the inter-person occlusion but also because of the necessity to estimate 3D poses in an absolute coordinate system (e.g., the camera coordinates), where each person is located properly to the other persons reflecting their locations in the real scenes \cite{Moon_2019_ICCV_3DMPPE,cheng2021graph}. 
However, the progress in 3D human pose estimation in the last few years mostly lies in single-person case. 
Existing methods can be generally grouped into top-down or bottom-up approaches, where the top-down methods employ a human detection  method to detect each person and then perform human pose estimation, while the bottom-up methods estimate all human keypoints simultaneously and then group them to form one or several skeletons. 

Multi-person pose estimation methods can be grouped into top-down and bottom-up approaches. Top-down methods first use human detection to detect every person in an image, and then process the cropped image patch of each detected person individually. The benefit of the top-down approach is that human detection can ensure that in each image patch there is only one target person. Moreover, the size of the image patches can be  normalized, alleviating the variation of human scales. 
The downside of the top-down approach is that, if human detection fails to detect one or a few persons, then there is no chance to predict the poses for those persons. 
Unlike top-down methods, bottom-up methods do not rely on human detection to detect each person, instead, they simultaneously detect all possible keypoints in a given image, and then group them to form individual human poses. 
The disadvantage of bottom-up methods is that the whole image is processed at once and no person-wise normalization can be performed, and thus the accuracy of pose estimation for small-scale persons particularly can be affected.

Based on the pros and cons of top-down and bottom-up methods, it is clear that neither one is suitable for all scenarios. Top-down methods can miss some persons in the case of occlusions, while bottom-up methods can not achieve good accuracy for small-scale persons. Such observation motivates us to develop a dual network for multi-person 3D pose estimation that integrates both the top-down and bottom-up networks, to robustly handle the challenging cases including occlusions and small-scale persons.

Table~\ref{tab:summary_td_bu} summarizes the top-down and bottom-up methods in 2D and 3D human pose estimation. The majority of them are top-down and bottom-up 2D human pose estimation, followed by methods in 3D human pose estimation  for single persons, which use the person-centric coordinates. Few methods are proposed to handle 3D multi-person pose estimation, particularly those that employ the bottom-up  approach.
Table~\ref{tab:summary} summarizes the differences between single-person and multi-person 3D pose estimation. Single-person methods use the person-centric coordinates, while multi-person methods use the camera-centric coordinates. This difference further influences the metrics and datasets used for evaluations.

\begin{figure}[t]
	\centering
	\includegraphics[width=\linewidth]{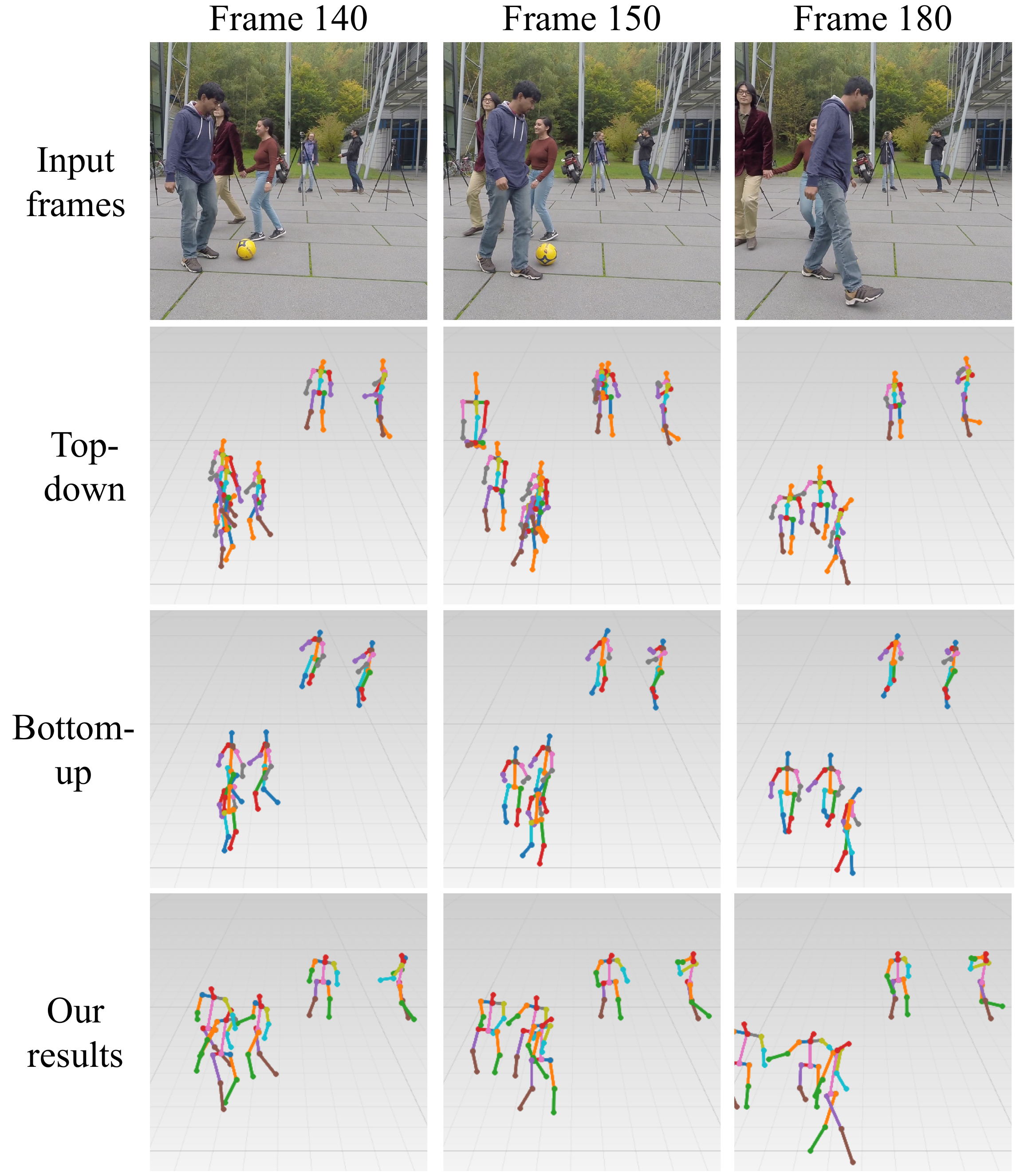}
	\vspace{-0.5em}
	\caption{Incorrect 3D multi-person pose estimation from existing top-down (2nd row) and bottom-up (3rd row) methods. The top-down method is RootNet~\cite{Moon_2019_ICCV_3DMPPE}, the bottom-up method is SMAP~\cite{zhen2020smap}. The input images are from MuPoTS-3D dataset~\cite{mehta2018single}. The top-down method suffers from inter-person occlusion and the bottom-up method is sensitive to scale variations (i.e., the 3D poses of the two persons in the back are inaccurately estimated). Our method substantially outperforms the state-of-the-art.}
	\label{fig:intriguing_example}
	\vspace{-1.0em}
\end{figure}

Existing top-down 3D pose estimation methods rely considerably on human detection to localize each person, prior to estimating the joints within the detected bounding boxes, e.g., \cite{pavllo20193d,cheng2019occlusion,Moon_2019_ICCV_3DMPPE}. These methods show promising performance for single-person 3D-pose estimation \cite{pavllo20193d,cheng2019occlusion}, yet since they treat each person individually, they have no awareness of non-target persons and the possible interactions. When multiple persons occlude each other, human detection also becomes unreliable. Moreover, when target persons are closely interacting with each other, the pose estimator may be misled by the nearby persons, e.g., predicted joints may come from the nearby non-target persons.

Recent bottom-up methods (e.g., \cite{benzine2020pandanet,fabbri2020compressed,zhen2020smap}) achieve comparable performance in multi-person datasets without using human detection.
As bottom-up methods consider multiple persons simultaneously and, in certain cases, may better distinguish the joints of different persons.
Unfortunately, without using detection, bottom-up methods suffer from the scale variations (i.e., no bounding box to normalize image patch), and the pose estimation accuracy is compromised, rendering inferior performance compared with  top-down approaches \cite{cheng2020higherhrnet}.
As shown in Fig. \ref{fig:intriguing_example}, neither top-down nor bottom-up approach alone can handle all the challenges at once, particularly the challenges of: inter-person occlusion, close interactions, and human-scale variations.

In this paper, we aim to integrate the top-down and bottom-up approaches to achieve more accurate and robust 3D multi-person pose estimation from a monocular video. To achieve this, we introduce a top-down network to estimate human joints inside each detected bounding box. 
Unlike existing top-down methods that only estimate one human pose given a bounding box, our top-down network predicts 3D poses for all persons inside the bounding box. 
Our top-down network generates the joint heatmaps and feeds them to our bottom-up network, which enables our bottom-up network to handle the scale variations.
Finally, the estimated 3D poses from both top-down and bottom-up networks are processed by our integration network to obtain the final estimated 3D poses given an image sequence.

Unlike existing methods' pose discriminators, which are designed solely for a single person, and consequently cannot enforce natural inter-person interactions, we propose a two-person pose discriminator that enforces two-person natural interactions. We also employ semi-supervised learning to mitigate the data scarcity problem where 3D ground-truth data is limited.
To address the domain gap problem between the training and testing data, we do optimization during test time. In particular, we propose novel approaches to refine the estimated 3D human poses through high-order temporal constraint, reprojection loss, and bone-length regularization.

This paper is based on our conference paper~\cite{cheng2021integrating}. Unlike our conference version, however, we add test time optimization to handle the gap between training and testing data in Section {\ref{sec:tto}}, which is critical for our method to process unseen videos. For this test-time optimization to work, we propose new unsupervised losses, i.e., high-order temporal constraints in Eq. {\ref{eq:highorder}}, reprojection loss in Eq. {\ref{eq:reproj}}, and bone-length regularization in Eq. {\ref{eq:bone_reg}}.
We also provide more analysis for inter-person pose discriminator in Table {\ref{tab:discrim}},  detailed information of the semi-supervised learning part in Fig. {\ref{fig:multi_persp}}, and more extensive qualitative comparisons against SOTA methods on MuPoTS and JTA datasets in Fig. {\ref{fig:mupots} and {\ref{fig:jta}}}. In summary, our contributions are listed as follows. 

\begin{itemize}
    \item We introduce a novel two-branch framework, where the top-down network detects multiple persons and the bottom-up network incorporates the normalized image patches in its process. Our framework gains benefits from the two networks, and at the same time, overcomes their shortcomings. 
	\item We employ multi-person pose estimation for our top-down network, which can effectively handle the inter-person occlusion and interactions caused by detection errors. 
	\item We incorporate human detection information into our bottom-up network so that  it can better handle the scale variation, which addresses the problem in existing bottom-up methods.
	\item Unlike the existing discriminators that focus on single person pose, we introduce a novel discriminator that enforces the validity of human poses of close pairwise interactions in the camera-centric coordinates.
	\item We propose high-order temporal constraint and bone length loss for test time optimization to improve the generalization capability of our method on testing videos.
\end{itemize}

\section{Related Works}

\noindent \textbf{Top-Down Monocular 3D Human Pose Estimation} 
Existing top-down 3D human pose estimation methods commonly use human detection as an essential part of their methods to estimate person-centric 3D human poses~\cite{martinez2017simple,nie2017monocular,mehta2017vnect,pavllo20193d,cheng2019occlusion,doersch2019sim2real,cheng2020sptaiotemporal}. 
They demonstrate promising performance on single-person evaluation datasets~\cite{human36ionescu,sigal2010humaneva}, unfortunately the performance decreases in multi-person scenarios, due to inter-person occlusion or close interactions \cite{mehta2017vnect,cheng2019occlusion}. Moreover, the produced person-centric 3D poses cannot be used for multi-person scenarios, where camera-centric 3D-pose estimation is needed.
Top-down methods process each person independently, leading to inadequate awareness of the existence of other persons nearby. 
As a result, they perform poorly on multi-person videos where inter-person occlusion and close interactions  are commonly present. 
Rogez et al.~\cite{rogez2017lcr,rogez2019lcr} develop a pose proposal network to generate bounding boxes and then perform pose estimation individually for each person. Recently, unlike previous methods that perform person-centric pose estimation, Moon et al.~\cite{Moon_2019_ICCV_3DMPPE} propose a top-down 3D multi-person pose-estimation method that can estimate the poses for all persons in an image in the camera-centric coordinates. However, the method still relies on detection and processes each person independently; hence it is likely to suffer from inter-person occlusion and close interactions.

\begin{figure*}[h]
    \centering
    \includegraphics[width=\textwidth]{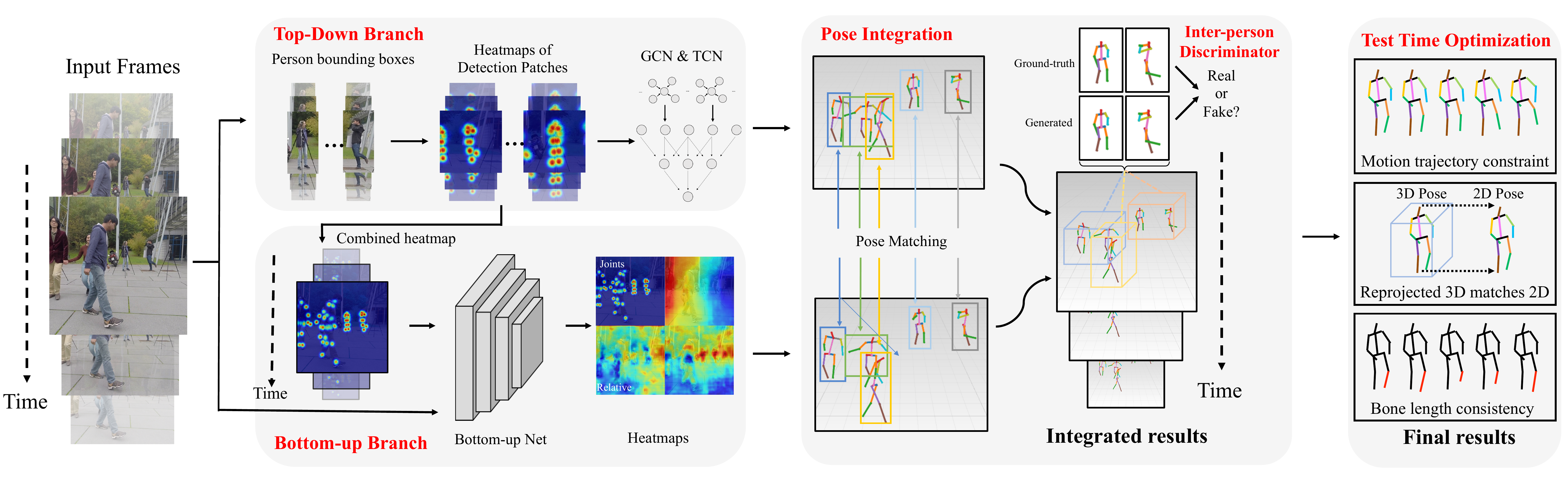}
    \caption{The overview of our framework. Our proposed method comprises four major components: 1) A top-down network to estimate fine-grained instance-wise 3D pose. 2) A bottom-up network to generate a global-aware camera-centric 3D pose. 3) An integration network to generate final estimation based on paired poses from top-down and bottom-up to take benefits from both networks. 4) A test time optimization process to  refine the obtained integrated 3D poses for the final result. Note that the semi-supervised learning part is a training strategy so it is not included in this figure.
    }
    \label{fig:pipeline}
\end{figure*}

\vspace{0.1cm}
\noindent \textbf{Bottom-Up Monocular 3D Human Pose Estimation} 
A few bottom-up methods have been proposed~\cite{fabbri2020compressed,zhen2020smap,mehta2020xnect,li2020hmor,lin2020hdnet}. Fabbri et al.~\cite{fabbri2020compressed} introduce an encoder-decoder framework to compress a heatmap first, and then decompress it back to the original representations in the test time for fast HD image processing. 
Mehta et al.~\cite{mehta2020xnect} propose to identify individual joints, compose full-body joints, and enforce temporal and kinematic constraints in three stages for real-time 3D motion capture. 
Li et al.~\cite{li2020hmor} develop an integrated method with lower computation complexity for human detection, person-centric pose estimation, and human depth estimation from an input image. 
Lin et al.~\cite{lin2020hdnet}  formulate the human depth regression as a bin index estimation problem for multi-person localization in the camera coordinate system.
Zhen et al.~\cite{zhen2020smap} estimate the 2.5D representation of body parts first and then reconstruct camera-centric multi-person 3D poses. 
These methods benefit from the nature of the bottom-up approach, which can process multiple persons simultaneously without relying on human detection.
However, since all persons are processed at the same scale, these methods are inevitably sensitive to human scale variations, which limits their applicability on wild videos. 

\vspace{0.1cm}
\noindent \textbf{Top-Down and Bottom-Up Combination} 
Earlier non-deep learning methods exploring the combination of top-down and bottom-up approaches for human pose estimation are in the forms of data-driven belief propagation, different classifiers for joint location and skeleton, or probabilistic Gaussian mixture modelling~\cite{hua2005learning,wang2010combined,kuo2011integration}. 
Recent deep learning based methods that attempt to make use of both top-down and bottom-up information are mainly on estimating 2D poses~\cite{hu2016bottom,tang2018deeply,cai2019exploiting,li2019multi}. 
Hu and Ramanan~\cite{hu2016bottom} propose a hierarchical rectified Gaussian model to incorporate top-down feedback with bottom-up convolutional neural network (CNN). 
Tang et al.~\cite{tang2018deeply} develop a framework with bottom-up inference followed by top-down refinement based on a compositional model of the human body.
Cai et al.~\cite{cai2019exploiting} introduce a spatial-temporal graph convolutional network (GCN) that uses both bottom-up and top-down features. 
These methods explore to benefit from top-down and bottom-up information. However, 
they are not suitable for 3D multi-person pose estimation because the fundamental weaknesses in both top-down and bottom-up methods are not addressed completely, which include inter-person occlusion caused detection and joints grouping errors, and the scale variation issue. 
Li et al.~\cite{li2019multi} adopt LSTM and combine bottom-up heatmaps with human detection for 2D multi-person pose estimation. They address occlusion and detection shift problems. Unfortunately, they use a bottom-up network and only add the detection bounding box as the top-down information to group the joints. Hence, their method is essentially still bottom-up and thus still vulnerable to human scale variations.

\vspace{0.1cm}
\noindent \textbf{Test Time Optimization for Human Pose Estimation} 
Although supervised learning approaches  show promising results in human pose estimation, it is unavoidable to encounter images/videos in testing which are not seen in training datasets in terms of appearance, motion,  pose, occlusions, etc. As a result, existing deep learning based pose estimation methods may not perform well on testing data. A few recent works explore to mitigate these issues \cite{zhang2020inference,yu2020multi,shimada2020physcap,su2021nerf}. Zhang et al.~\cite{zhang2020inference} propose to check the validity of estimated 2D pose and the consistency of lifted 3D poses from randomly projected 2D poses to refine the estimated 3D human pose. Cheng et al.~\cite{yu2020multi} propose to use 2D pose estimator's confidence to weight a re-projection loss in inference stage to make pose corrections when estimated 3D pose is erroneous but the 2D pose is more accurate where 2D pose estimator is trained on dataset with larger variations in appearance and pose (i.e., 2D annotation is easier to obtain compared to 3D ground-truth). Shimada et al.~\cite{shimada2020physcap} propose to enforce physical constraints to ensure the estimated 3D human poses are physically plausible. Su et al.~\cite{su2021nerf} propose a neural radiance fields (NeRF) based 3D pose correction framework where estimated 3D pose of a person is used as input for a customized NeRF to render the person to compute a image difference loss against input image with the person to correct the initial 3D pose estimated.

\section{Method}

Fig. \ref{fig:pipeline} shows our pipeline, which consists of four major parts to accomplish the multi-person camera-centric 3D human pose estimation: a top-down network for fine-grained instance-wise pose estimation, a bottom-up network for global-aware pose estimation, an integration network to integrate the estimation results of the top-down and bottom-up networks with inter-person pose discriminator, and a test time optimization process to refine and obtain the final 3D human poses. Moreover, a semi-supervised training process is proposed to enhance the 3D pose estimation based on reprojection consistency. 

\subsection{Top-Down Network}
\label{sec:top-down}

\begin{figure}
    \centering
    \includegraphics[width=0.9\linewidth]{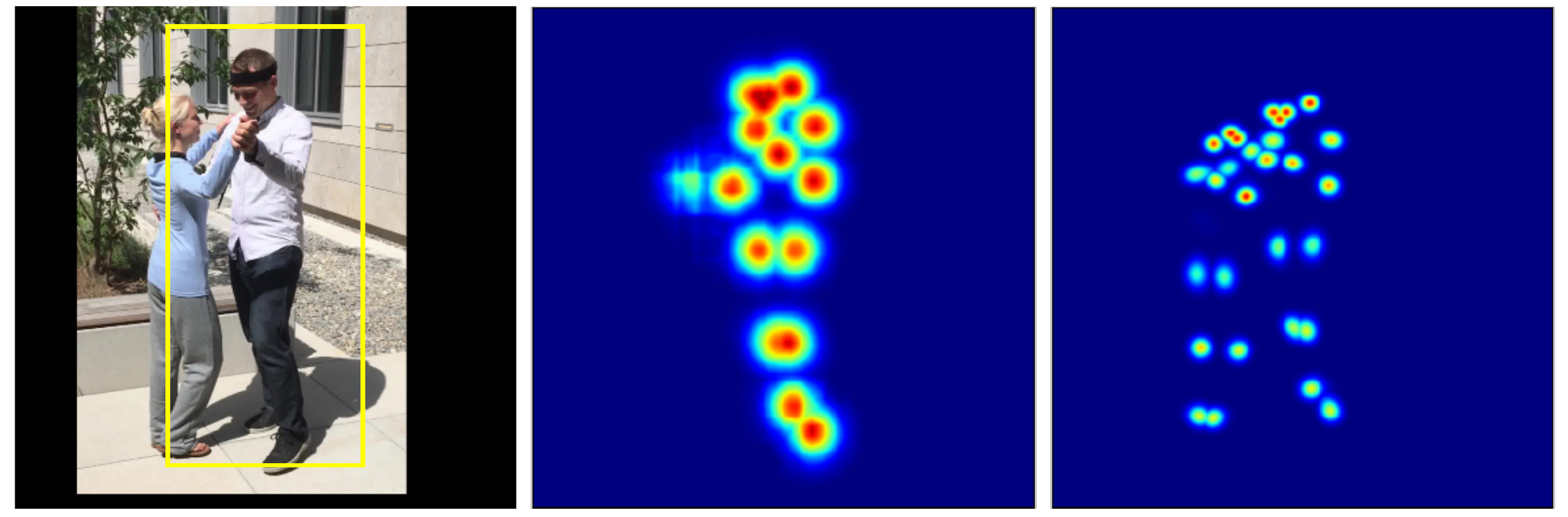}
    \caption{Examples of estimated heatmaps of human joints. 
    The left image shows the input frame overlaid with an inaccurate detection bounding box (i.e., only one person detected).
    The middle image shows the estimated heatmap of existing top-down methods. The right image shows the heatmap of our top-down network.
    }
    \label{fig:topdownfail}
\end{figure}

Given a human detection bounding box, existing top-down methods  estimate full-body joints of one person. 
Consequently, if there are multiple persons inside the box or partially out-of-bounding box body parts, the full-body joint estimation are likely to be erroneous.
Fig. \ref{fig:topdownfail} shows such failure examples of existing methods. 
In contrast, our method produces the heatmaps for all joints inside the bounding box (i.e., enlarged to accommodate inaccurate detection), and estimates the ID for each joint to group them into corresponding persons, similar to~\cite{newell2017associative}. 

Given an input video, for every frame we apply a human detector \cite{he2017mask}, and crop the image patches based on the detected bounding boxes.
A 2D pose detector \cite{cheng2020higherhrnet} is applied to each patch to generate heatmaps for all human joints, such as shoulder, pelvis, ankle, and etc. 
Specifically, our top-down loss of 2D pose heatmap is an L2 loss between the predicted and ground-truth heatmaps, formulated as:
\begin{equation}
    L^{TD}_{hmap} = |H - \Tilde{H}|^2_2, 
\label{eq:hmap}
\end{equation}
where $H$ and $\Tilde{H}$ are the predicted and ground-truth heatmaps, respectively. 

Having obtained the 2D pose heatmaps, a directed GCN network is used to refine the potentially incomplete poses caused by occlusions or partially out-of-bounding box body parts, and two TCNs are used to estimate both person-centric 3D pose and camera-centric root depth based on a given sequence of 2D poses similar to~\cite{cheng2021graph}. As the TCN requires the input sequence of the same instance, a pose tracker \cite{umer2020self} is used to track each instance in the input video. We also apply data augmentation in training our TCN so that it can handle occlusions \cite{cheng2019occlusion}. 

\subsection{Bottom-Up Network}
\label{sec:bottom-up}

Top-down methods perform estimation inside the bounding boxes, and thus are lack of global awareness of other persons, leading to difficulties to estimate poses in the camera-centric coordinates. 
To address this problem, we further propose a bottom-up network that processes multiple persons simultaneously.
Since the bottom-up pose estimation suffers from human scale variations, we concatenate the heatmaps from our top-down network with the original input frame as the input of our bottom-up network.
With the guidance of the top-down heatmaps, which are the results of the object detector and pose estimation based on  the normalized boxes, the estimation of the bottom-up network will be more robust to scale variations.
Our bottom-up network outputs four heatmaps : a 2D pose heatmap, ID-tag map, relative depth map, and root depth map. 
The 2D pose heatmap and ID-tag map are defined in the same way as in the previous section (\ref{sec:top-down}). 
The relative depth map refers to the depth map of each joint with respect to its root (pelvis) joint. 
The root depth map represents the depth map of the root joint. 

In particular, the loss functions $L^{BU}_{hmap}$ and $L^{BU}_{id}$ for the heatmap and ID-tag map are similar to \cite{newell2017associative}. 
In addition, we apply the depth loss to the estimations of both the relative depth map $h^{rel}$ and the root depth $h^{root}$. Please see Fig. \ref{fig:pipeline} for example of the four estimated heatmaps from the bottom-up network. For $N$ persons and $K$ joints, the loss can be formulated as:
\begin{equation}
    L_{depth} = \frac{1}{NK}\sum_n \sum_k |h_{k}(x_{nk}, y_{nk}) - d_{nk}|^2,
\end{equation}
where $h$ is the depth map and $d$ is the ground-truth depth value. 
Note that, for the pelvis (i.e., the root joint), the depth is a camera-centric depth.
For other joints, the depth is relative with respect to the corresponding root joint. 

We group the heatmaps into instances (i.e., persons), and retrieve the joint locations using the same procedure as in the top-down network. 
Moreover, the values of the camera-centric depth of the root joint $z^{root}$ and the relative depth for the other joints $z^{rel}_k$ are obtained by retrieving from the corresponding depth maps where the joints (i.e., root or others) are located.
Specifically:
\begin{eqnarray}
    z^{root}_{i} &=& h^{root}(x^{root}_i, y^{root}_i) \\
    z^{rel}_{i,k} &=& h^{rel}_k (x_{i,k}, y_{i,k})
\end{eqnarray}
where $i,k$ refer to the $i_{th}$ instance and $k_{th}$ joint, respectively.

\subsection{Integration  with Interaction-Aware Discriminator}
\label{sec:integ}

Having obtained the results from the top-down and bottom-up networks, 
we first need to find the corresponding poses between the results from the two networks, i.e., the top-down pose $P^{TD}_i$ and bottom-up pose $P^{BU}_j$ belong to the same person. Note that $P$ stands for camera-centric 3D pose throughout this paper. 

Given two pose sets from bottom-up network $P^{BU}$ and top-down network $P^{TD}$, we match the poses from both sets, in order to form pose pairs. 
The similarity of two poses is defined as:
\begin{eqnarray}
    {\rm Sim}_{i,j} = \sum_{k=0}^K \min(c^{BU}_{i,k}, c^{TD}_{j,k}) {\rm OKS} (P^{BU}_{i,k}, P^{TD}_{j,k}),
\end{eqnarray}
where:
\begin{eqnarray}
    {\rm OKS}(x,y) = \exp(- \frac{d(x,y)^2}{2s^2\sigma^2}),
\end{eqnarray}
${\rm OKS}$ stands for object keypoint similarity \cite{xiao2018simple}, which measures the joint similarity of a given joint pair. 
$d(x,y)$ is the Euclidean distance between two joints. 
$s$ and $\sigma$ are two controlling parameters. 
${\rm Sim}_{i,j}$ measures the similarity between the $i_{th}$ 3D pose $P^{BU}_{i}$ from the bottom-up network and the $j_{th}$ 3D pose $P^{TD}_{j}$ from the top-down network  over $K$ joints.
Note that both poses from top-down $P^{TD}$ and bottom-up $P^{BU}$ are camera-centric; thus, the similarity is measured based on the camera coordinate system. 
The $c^{BU}_{i,k}$ and $c^{TD}_{j,k}$ are the confidence values of joint $k$ for 3D poses $P^{BU}_{i}$ and $P^{TD}_{j}$, respectively. 
Having computed the similarity matrix between the two sets of poses $P^{TD}$ and $P^{BU}$ according to the ${\rm Sim}_{i,j}$ definition, the Hungarian algorithm \cite{kuhn1955hungarian} is used to obtain the matching results.

Once the matched pairs are obtained, we feed each pair of the 3D poses and the confidence score of each joint to our integration network.
Our integration network consists of 3 fully connected layers, which outputs the final estimation. 

\vspace{0.3cm}
\noindent \textbf{Integration Network Training}
To train the integration network, we take some samples from the ground-truth 3D poses.
We apply data augmentation: 
1) random masking the joints with a binary mask $M^{kpt}$ to simulate occlusions; 
2) random shifting the joints to simulate the inaccurate pose detection; 
and  3) random zeroing one from a pose pair to simulate unpaired poses. 
The loss of the integration network is an L2 loss between the predicted 3D pose and its ground-truth:
\begin{equation}
    L_{int} = \frac{1}{K} \sum_k |P_k - \Tilde{P}_k|^2, 
\end{equation}
where $K$ is the number of the estimated joints. $P$ and $\Tilde{P}$ are the estimated and ground-truth 3D poses, respectively.

\vspace{0.3cm}
\noindent \textbf{Inter-Person Discriminator}
For training the integration network, we propose a novel inter-person discriminator.
Unlike most existing discriminators  for human pose estimation (e.g. \cite{wandt2019repnet,cheng2020sptaiotemporal}), where they can only discriminate the plausible 3D poses of one person, we propose an interaction-aware discriminator to enforce the interaction of a pose pair is natural and reasonable, which not only includes the existing single-person discriminator, but also generalize to interacting persons. 
Specifically, our discriminator contains two sub-networks: $D_1$, which is dedicated for  one person-centric 3D poses; and, $D_2$, which is dedicated for a pair of camera-centric 3D poses from two persons. 
We apply the following loss to train the network, which is formulated as:
\begin{equation}
L_{dis} = log(\Tilde{C}) + log(1-C)
\end{equation}
where:
\begin{equation}
\begin{split}
    &C = 0.25 (D_1(P^a) + D_1(P^b)) +  0.5 D_2(P^a, P^b) \\
    &\Tilde{C} = 0.25 (D_1(\Tilde{P}^a) + D_1(\Tilde{P^b})) + 0.5 D_2(\Tilde{P}^a, \Tilde{P^b})
\end{split}
\end{equation}
where $P^a, P^b$ are the estimated poses of person $a$ and person $b$, respectively. $\Tilde{P}$ are the estimated and ground-truth 3D poses, respectively. 

\subsection{Semi-Supervised Training}
\label{sec:ssl}

Semi-supervised learning is an effective technique to improve the network performance, particularly when the data with ground-truths are limited.
A few works also explore how to make use of the unlabeled data~\cite{chen2019unsupervised,umer2020self,xu20203dssl}. 
%
%
We first train our network with the 3D ground-truth dataset only, and then use the trained network to generate the pseudo-labels of unlabelled data, which are then used to fine tune the network. 

\begin{figure}
    \centering
    \includegraphics[width=\linewidth]{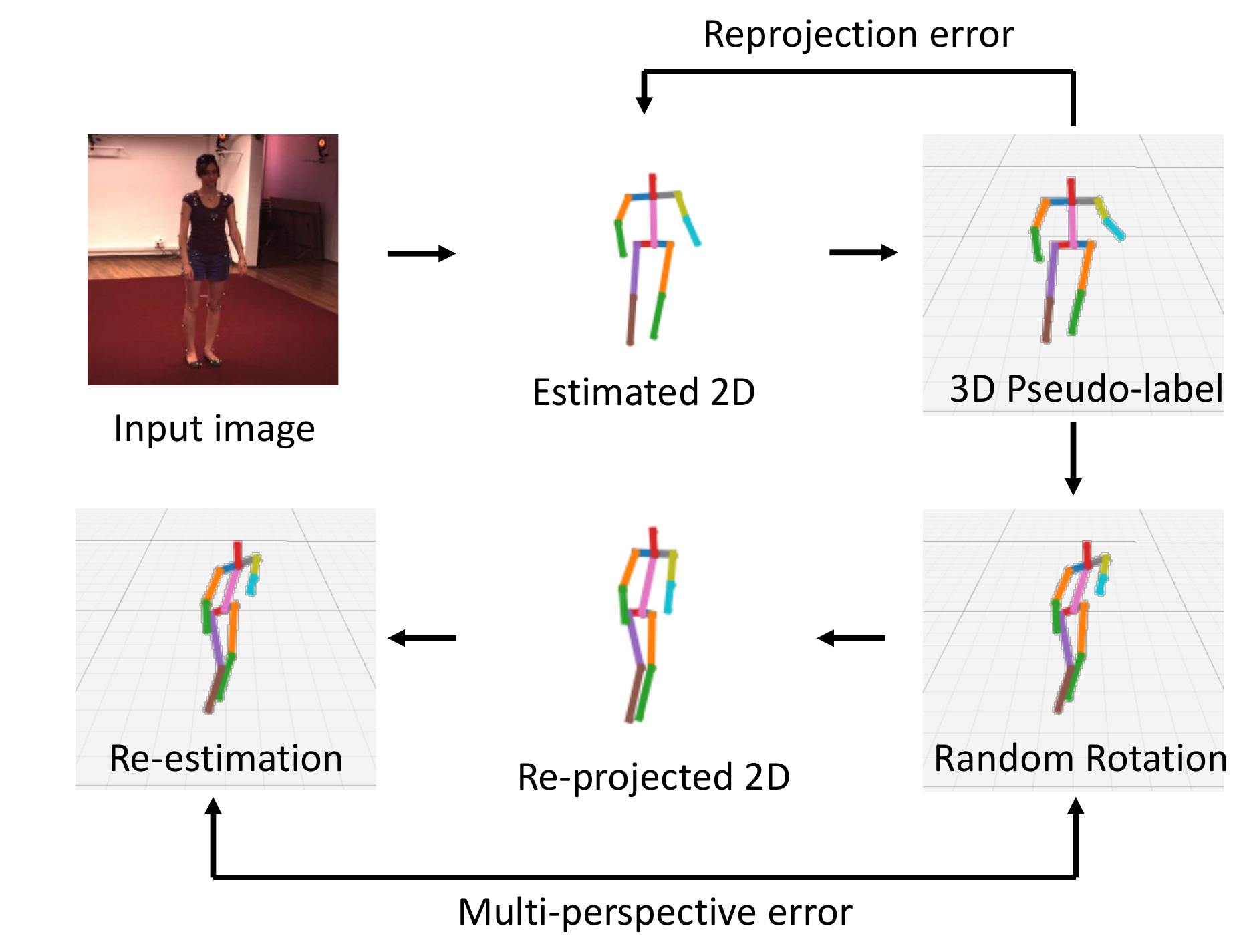}
    \caption{The illustration of our SSL pipeline. The SSL aims to keep two consistency: reprojection and multi-perspective.}
    \label{fig:sslpipeline}
\end{figure}

The generated pseudo-labels cannot be directly used because some of them are likely incorrect. 
Unlike recent noisy student training strategy~\cite{xie2020self}, where data with ground-truth labels and pseudo-labels are mixed to train the student network by adding various types of noise (i. e., augmentations, dropout, etc), we use reprojection loss and multi-perspective loss to correct the errors in the pseudo-labels.
Our Semi-supervised Learning (SSL) pipeline is shown in Fig. \ref{fig:sslpipeline}. First, we use the trained model to generate the pseudo-label of the unlabelled data, which is the COCO dataset in our experiment. Note that, we use only the images, and not the 2D ground-truths of the joints to mimic the unlabelled data scenario. 
Therefore, we use two consistency terms to measure the quality of all the pseudo-labels: the reprojection loss and multi-perspective loss. 

As the pose variations of 2D datasets are more abundant than those of 3D datasets, e.g. COCO compared to H36M, the estimated 2D poses are more robust than the estimated 3D poses in terms of different environments and poses. Existing reprojection loss \cite{wandt2019repnet} measures the deviation between generated 3D poses and detected 2D poses. Unlike this, we make use of the confidence of the joints from the 2D pose heatmap as weight in computating  the reprojection loss to adjust adaptively how much we should enforce the reprojected 3D poses to match the estimated 2D poses based on the confidence of the joints. Thus, the reprojection loss is formulated as:
\begin{equation}
    L_{rep} = \frac{1}{K} \sum^{K}_{k=1} C_k |rep(X_{3D,k}) - X_{2D,k}|^2
\label{eq:reproj}
\end{equation}
where the $X_{3D}$ is the predicted 3D pose from the network, and $X_{2D}$ stands for the 2D estimations from our multi-person 2D pose estimator. $rep(\cdot)$ is the reprojection function from 3D to 2D. $K$ stands for the number of joints in total. Moreover, the loss is a weighted sum of each joint's confidence score $C_k$, which is the maximum value of the joint's heatmap.

A multi-perspective loss is used as an additional measure to enforce the consistency of the predicted 3D poses from different viewing angles \cite{chen2019unsupervised}. As shown in Fig. \ref{fig:sslpipeline}, given a pseudo-label 3D pose $P^{pse}_{3D}$, we randomly rotate the pose along $y$ axis (i.e., y-axis is perpendicular to the ground plane) to obtain $P'^{pse}_{3D}$, and re-project it to the 2D coordinates. Finally, we predict the $P''^{pse}_{3D}$ based on the 2D projection from $P_{3D}^{pses}$.

The reprojection loss measures the deviation between the projection of generated 3D poses and the detected 2D poses. Since there are more abundant data variations in 2D pose dataset compared to 3D pose dataset (e.g., COCO is much larger compared to H36M), the 2D estimator is expected to be more reliable than its 3D counterpart. 
Therefore, minimizing a reprojection loss is helpful to improve the accuracy of 3D pose estimation.

The multi-perspective loss, $L_{mp}$, measures the consistency of the predicted 3D poses from different viewing angles. This loss indicates the reliability of the predicted 3D poses. Based on the two terms, our semi-supervised loss, $L_{\rm SSL}$, is formulated as,
\begin{equation}
    L_{\rm SSL} = w(L_{rep} + L_{mp}) + L_{dis},
\end{equation}
where $w$ is a weighting factor to balance the contribution of the reprojection and multi-perspective losses. 
In the training stage, $w$ first focuses on easy samples and gradually includes the hard samples. The weight, $w$, is  formulated as:
\begin{equation}
    w = {\rm softmax}(\frac{E_{rep}}{r}) + {\rm softmax}(\frac{E_{mp}}{r}),
\end{equation}
where $r$ is the number of training epochs.


\subsection{Test Time Optimization}
\label{sec:tto}

Unlike the semi-supervised training in Section \ref{sec:ssl}, where the network is fine-tuned with the extra data. In this section, the purpose of test time optimization is to refine the estimated 3D poses at the inference time without permanently updating the network itself. 
As the training process focuses on fitting an optimal model only on the training set, and is not aware of the gap between training and testing data, test time optimization brings performance improvement by adopting the  prediction from trained model to the testing data. 

The domain gap can be observed from both temporal and spatial perspectives. First, the speed of motion varies from video to video. In training data, the speed of motion is fixed and cannot cover all possible variations even with augmentations, which results in the gap from testing data. Since the TCN takes temporal information into consideration, the change of motion speed can drag down the accuracy of predicted 3D poses. Second, since there is a limited variation of bone lengths in the training data, the TCN learns a strong prior of fixed bone length. This will result in inaccurate predictions because the same 2D poses can correspond to different 3D poses with different bone lengths. To this end, we propose two strategies for inference stage optimization with two regularizations: trajectory, reprojection, and bone-length regularization.

First, we introduce the trajectory regularization to restrict the motion from unusual motions. Different motions follow a certain trajectory which can be expressed as a motion function $f_m(t)$. Taking a Taylor expansion of this function, we can approximate the motion function by a high-order polynomial. Our target is to predict the pose at the $t_{th}$ future frame given the poses in frames $0$ to $t-1$, which is formulated as:
\begin{equation}
    f_m(t) = c_0 + c_1 t + c_2 t^2 + c_3 t^3 + ... .
\end{equation}
As an approximation, we take the first 3 orders and approximate the coefficients $c_0, c_1, c_2, c_3$ using linear regression. The temporal trajectory constraint is the mean squared error between the estimated pose $P_{t}$ and the trajectory $f_m(t)$ at frame $t$:
\begin{equation}
    L_{traj} = (P_{t} - f_m(t))^2.
    \label{eq:highorder}
\end{equation}

Second, we propose to utilize the reprojection regularization to make the predicted 3D joints to be consistent with 2D joints. The same function as Eq.~(\ref{eq:reproj}) is used as the reprojection loss.
Third, as the bone length should be consistent for the same person across all frames in one video, we propose to use a bone length constraint to regularize the variation of each estimated 3D pose sequence. We first set up latent variables ${B_0, B_1, ...}$ to represent the lengths of all bones in human skeleton, and the bone loss is represented as:
\begin{equation}
    L_{bone} = \sum_{t} \sum_{i} (\Tilde{B_{i,t}} - B_{i,t})^2,
    \label{eq:bone_reg}
\end{equation}
where $\Tilde{B_i}$ is the bone length of estimated poses. By minimizing the equation above, we concurrently minimize the variance of bone lengths across the whole video, as well as the latent variables. 

In summary, the loss function for the test time optimization is:
\begin{equation}
    L_{TTO} = L_{traj} + c_{rep}L_{rep} + c_{bone}L_{bone},
\end{equation}
where $c_{rep}$ and $c_{bone}$ are the coefficients for reprojection and bone losses. Empirically, we observe that using the two-stage training strategy with different coefficients between different losses will result in better convergence of the test-time optimization process. First, we train with $c_{rep}=0.1$ and $c_{bone}=1$ with 3000 iterations. Then, we increase the $c_{rep}$ to $100$ to restrict more on the reprojection consistency for 3000 more iterations. We refer to the first step as "one-stage" and to the whole process as "two-stage" optimization strategy.

\section{Experiments}

\subsection{Datasets}
\label{sec:datasets}

We use MuPoTS-3D~\cite{mehta2018single} and JTA~\cite{fabbri2018learning} datasets to evaluate the camera-centric 3D multi-person pose estimation performance by following the existing methods~\cite{Moon_2019_ICCV_3DMPPE,fabbri2020compressed} and their training protocols (i.e., train, test split). In addition, we use 3DPW~\cite{3DPW} to evaluate person-centric 3D multi-person pose estimation performance following~\cite{humanMotionKanazawa19,sun2019human}. We also perform evaluation on the widely used Human3.6M dataset~\cite{human36ionescu} for person-centric 3D human pose estimation following~\cite{pavllo20193d,wandt2019repnet}. 


\textbf{MuPoTS-3D}~\cite{mehta2018single} is a 3D multi-person testing set that consists of $>$8000 frames of 5 indoor and 15 outdoor scenes, and its corresponding training set is augmented from 3DHP, called MuCo-3DHP. The ground-truth 3D pose of each person in a video is obtained from a multi-view markerless motion capture system, which is suitable for evaluating 3D multi-person pose estimation performance in both person-centric and camera-centric coordinates.
Following~\cite{Moon_2019_ICCV_3DMPPE}, the training set (MuCo-3DHP) is used for training our bottom-up network, and MuPoTS-3D is used only for performance evaluation.

\textbf{JTA}~\cite{fabbri2018learning} is a synthesized dataset from Grand Theft Auto V (GTA-V) game scene including various illumination, viewpoints, and occlusion. It is a multi-person dataset with at most $32$ persons appearing in one frame. In addition, the images also demonstrate large person size variation as the crowd spread from close to the camera and far from the camera in various scenes. Because of these reasons, even though it is a synthetic dataset, we want to evaluate it. The dataset contains 512 videos, in which there are 256, 128, 128 for training, validation and testing, respectively. We follow the work \cite{fabbri2020compressed} to estimate the F1 score under different distance thresholds as a performance evaluation metric. 


\textbf{Human3.6M}~\cite{human36ionescu} is widely used for 3D human pose estimation. The dataset contains 3.6 million single-person images where an actor performs different activities in mocap studio at each video clip, so it is suitable for evaluation of 3D single-person pose estimation. Therefore, Human3.6M is used for person-centric pose estimation evaluation to demonstrate the performance of the proposed method against other person-centric human pose estimation methods. Following previous works~\cite{hossain2018exploiting,pavllo20193d,wandt2019repnet}, the subject 1,5,6,7,8 are used for training, and 9 and 11 for testing.


\begin{figure}[t]
    \centering
    \includegraphics[width=0.9\linewidth]{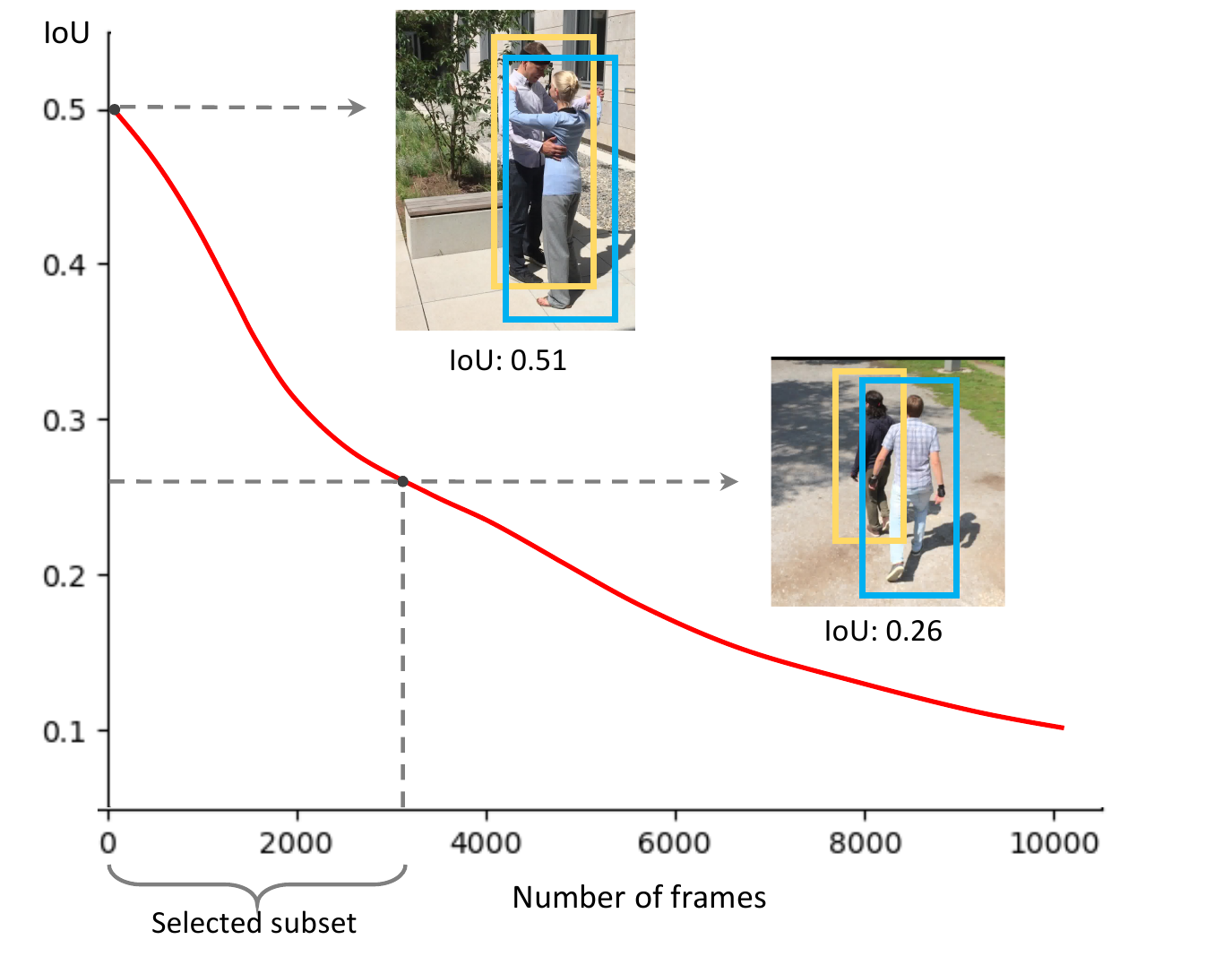}
    \caption{Interaction IoUs of 3DPW test set. 
    }
    \label{fig:iou}
\end{figure}

\textbf{3DPW}~\cite{3DPW} is an outdoor multi-person video dataset for 3D human pose reconstruction. Following previous methods~\cite{humanMotionKanazawa19,sun2019human}, we use 3DPW for testing without any fine-tuning. 
The ground-truth of 3DPW is SMPL 3D mesh model~\cite{loper2015smpl}, where the definition of joints differs from what is used in 3D human pose estimation (skeleton-based) like Human3.6M, so 3DPW is rarely used in the evaluation of skeleton-based methods~\cite{tripathi2020posenet3d}. 

As evaluation on 3DPW cannot objectively reflect the performance of the skeleton-based methods, due to different definitions of joints. We select the top 3000 frames with the largest IoU between the target person (i.e., the person with 3D ground-truth label) and other persons based on detection out of 3DPW test set to create an inter-person occlusion subset, and then perform evaluation on it. The IoU statistics of the 3DPW test set is shown in Fig. \ref{fig:iou}, and the threshold at $3000^{th}$ frame is $0.26$. 

In fact, the error on this subset is still not a good performance indicator, the performance change of a method between the full testing set and this subset can measure how well the method can handle the inter-person occlusion problem. As shown in Table \ref{tab:3dpw}, our method shows the smallest error increase among all the existing methods, which demonstrates that our method is indeed capable of handling inter-person occlusion more effectively.

\subsection{Implementation Details}

We use HRNet-w32 \cite{sun2019hrnet} as the backbone network for both multi-person pose estimators in the top-down and bottom-up networks. The top-down network is trained for $100$ epochs on the COCO dataset \cite{lin2014microsoft} with the Adam optimizer and learning rate $0.001$. The bottom-up network is trained for $50$ epochs with the Adam optimizer and learning rate $0.001$ on a combined dataset of MuCO \cite{singleshotmultiperson2018} and COCO \cite{lin2014microsoft}. 

\subsection{Evaluation Metrics}

Since the majority of 3D human pose estimation methods produce person-centric 3D poses, to be able to compare, we perform person-centric 3D human pose estimation. We use Mean Per Joint Position Error (MPJPE), Procrustes analysis MPJPE (PA-MPJPE), Percentage of Correct 3D Keypoints (PCK), and area under PCK curve from various thresholds ($AUC_{rel}$) following the literature~\cite{Moon_2019_ICCV_3DMPPE,pavllo20193d,cheng2020sptaiotemporal}. Since we focus on 3D multi-person camera-centric pose estimation, we also use the metrics designed for evaluating performance in the camera coordinate system, including average precision of 3D human root location ($AP_{25}^{root}$) and PCK$_{abs}$, which is PCK without root alignment to evaluate the absolute camera-centric coordinates from~\cite{Moon_2019_ICCV_3DMPPE}, and F1 value following~\cite{fabbri2020compressed}. 

\subsection{Ablation Studies}

\begin{table}[t]
	\footnotesize
	\centering
	\begin{tabular}{c|c|c|c|c}
		\cline{1-5}
		\cline{1-5}
		\rule{0pt}{2.6ex}
		\textbf{Method} & $AP_{25}^{root}$ & $AUC_{rel}$ & PCK & PCK$_{abs}$\\
		\cline{1-5}
		\rule{0pt}{2.6ex}
		TD (w/o MP) & 43.7 & 41.0 & 81.6 & 42.8\\
		TD (w MP) & 45.2 & 48.9 & 87.5 & 45.7\\
		BU (w/o CH) & 44.2 & 34.5 & 76.6 & 40.2\\
		BU (w CH) & 46.1 & 35.1 & 78.0 & 41.5\\
		TD + BU (w/o MP, CH) & 44.9 & 42.6 & 82.8 & 43.1\\
		TD + BU (hard) & 46.1 & 48.9 & 87.5 & 46.2\\
		TD + BU (linear) & 46.1 & 49.2 & 88.0 & 46.7\\
		TD + BU (w/o PM) & 46.0 & 48.6 & 85.5 & 45.3 \\
		TD + BU (IN) & \underline{46.3} & 49.6 & 88.9 & 47.4\\
		TD + BU (IN) + SSL & \underline{46.3} & \underline{50.6} & \underline{89.3} & \underline{47.7} \\
		TD + BU (IN) + TTO & \textbf{46.5} & 50.1 & {89.2} & 47.6 \\
		Full model & \textbf{46.5} & \textbf{50.7} & \textbf{89.6} & \textbf{48.1} \\
		\cline{1-5}
	\end{tabular}
	\vspace{0.5em}
	\caption{Ablation study on MuPoTS-3D dataset. TD, BU, MP, CH, IN, PM, SSL, TTO stand for top-down, bottom-up, multi-person pose estimator, combined heatmap, integration network, pose matching, semi-supervised learning, and test time optimization, respectively. Best in \textbf{bold}, second best \underline{underlined}. }
	\label{tab:Ablation}
\end{table}

\noindent \textbf{Analysis of the Major Components} Ablation studies are performed to validate the effectiveness of each sub-module of our framework as in Table~\ref{tab:Ablation}. 
We validate our top-down network by using an existing top-down pose estimator (i.e.,  detection of one full-body joints) as a baseline, abbreviated as TD (w/o MP) to compare to our top-down network denoted as TD (w MP). 
We also validate our bottom-up network by using existing bottom-up heatmap estimation (i.e., estimate persons at the same scale) as a baseline, named BU (w/o CH) to compare to our bottom-up network, called BU (w CH). 
To evaluate our integration network, we use three baselines. 
The first is a straightforward integration by combining existing TD and BU networks, called TD + BU (w/o MP, CH). 
The second is hard integration, abbreviated TD + BU (hard), where the top-down person-centric pose is always used,  plus the root depth from the bottom-up network. 
The third is linear integration, abbreviated TD + BU (linear), where the person-centric 3D pose from top-down is combined with its corresponding bottom-up one based on the confidence values of the estimated heatmap.
Compared against these baselines, the proposed integration solution is abbreviated as TD + BU (IN), we also provide the result where pose matching is not used, abbreviated TD + BU (w/o PM) to show the effect of pose matching. 
To evaluate other major components in our method, we also provide the following baselines: TD + BU (IN) + SSL where semi-supervised learning is used; TD + BU (IN) + TTO where test time optimization is used but no SSL. 
In the last row of Table~\ref{tab:Ablation}, it is the result of our full model, which is TD + BU (IN) + SSL + TTO.

\begin{table}[t]
	\footnotesize
	\centering
	\begin{tabular}{c|c|c|c|c}
		\cline{1-5}
		\cline{1-5}
		\rule{0pt}{2.6ex}
		\textbf{Method} & $AP_{25}^{root}$ & $AUC_{rel}$ & PCK & PCK$_{abs}$\\
		\cline{1-5}
		\rule{0pt}{2ex}
		Rep & \textbf{46.3} & 43.4 & 77.2 & 40.7\\
		MP & \textbf{46.3} & 32.2 & 72.8 & 29.5\\
		Rep+dis & \textbf{46.3} & \underline{49.9} & \underline{89.1} & \underline{46.8}\\
        Rep+MP+dis & \textbf{46.3} & \textbf{50.6} & \textbf{89.3} & \textbf{47.7}\\
		\cline{1-5}
	\end{tabular}
	\vspace{0.5em}
	\caption{Ablation study on MuPoTS-3D dataset. Rep, MP, and dis stand for reprojection, multi-perspective, and discriminator. Best in \textbf{bold}, second best \underline{underlined}.}
	\label{tab:ssl}
	\vspace{-0.5em}
\end{table}

\begin{table}[]
    \centering
    \begin{tabular}{c|c|c|c|c}
    \cline{1-5}
    \cline{1-5}
    \rule{0pt}{2.6ex}
        \textbf{Method} & $AP_{25}^{root}$ & $AUC_{rel}$ & PCK & PCK$_{abs}$\\
        \cline{1-5}
        \rule{0pt}{2ex}
        Baseline & 46.3 & 43.4 & 77.2 & 40.7\\
        Dis1 & 46.1 & 48.9 & 87.3 & 46.1 \\
        Dis2 & \textbf{46.3} & \textbf{49.9} & \textbf{89.1} & \textbf{46.8} \\
        Dis3 & 46.3 & 49.7 & 89.0 & 46.7 \\
        \cline{1-5}
    \end{tabular}
    \vspace{0.5em}
    \caption{Parameter analysis of different number of poses to check in the inter-person discriminator on MuPoTS-3D dataset. Dis1, Dis2, and Dis3 stand for checking one, two, and three poses in the inter-person discriminator. Best in \textbf{bold}.}
    \label{tab:discrim}
    \vspace{-0.5em}
\end{table}

\begin{table}[t]
    \centering
    \begin{tabular}{c|c|c}
    \cline{1-3}
    \cline{1-3}
    \rule{0pt}{2.6ex}
        \textbf{Method} & MPJPE (w. GT) & MPJPE (w/o GT)\\ 
        \cline{1-3}
        \rule{0pt}{2ex}
        Baseline & 44.84 & 58.71\\
        + reprojection regularization & 39.27 & 54.45\\
        + trajectory regularization &38.92 & 53.91\\
        + bone length consistency &37.11 & 51.30\\
        + one-stage optimization & \underline{35.75} & \underline{50.66}\\
        + two-stage optimization & \textbf{34.95} & \textbf{49.31}\\
        \cline{1-3}
    \end{tabular}
    \vspace{0.5em}
    \caption{Ablation study of test time optimization on Human3.6M dataset. MPJPE (w. GT) means 2D ground truth is used, MPJPE (w/o GT) means 2D pose estimation is used instead of the ground truth. Best in \textbf{bold}, second best \underline{underlined}. }
    \label{tab:tto}
    \vspace{-0.5em}
\end{table}

\begin{table}
    \centering
    \begin{tabular}{c|c|c|c}
    \cline{1-4}
    \cline{1-4}
    \rule{0pt}{2.6ex}
         \textbf{Order 1} & \textbf{Order 2} & \textbf{Order 3} & MPJPE  \\  
         \cline{1-4}\rule{0pt}{2ex}2 & 0 & 0 & 9.18 \\
         3 & 0 & 0 & 9.24 \\
         5 & 0 & 0 & 10.51 \\ \cline{1-4}
         2 & 3 & 3 & 5.88 \\
         \textbf{2} & \textbf{5} & \textbf{5} & \textbf{5.84} \\
         2 & 7 & 7 & 5.93 \\
         3 & 3 & 3 & 5.94 \\
         3 & 5 & 5 & 7.21 \\
         3 & 7 & 7 & 7.71 \\ \cline{1-4}
    \end{tabular}
    \vspace{0.5em}
    \caption{Ablation study on the number of temporal window lengths for different motion trajectory order using ground-truth 3D poses on Human3.6M dataset. The error is measured by MPJPE. Best in \textbf{bold}.s}
    \label{tab:traj}
    \vspace{-0.5em}
\end{table}

As shown in Table~\ref{tab:Ablation}, we observe that our top-down network, bottom-up network, and integration network clearly outperform their corresponding baselines. 
Our top-down network tends to have better person-centric 3D pose estimation results compared with our bottom-up network, because the top-down network benefits from not only the multi-person pose estimator, but also the GCN and TCN that help to deal with inter-occluded poses.
On the contrary, our bottom-up network achieves better performance for the root joint estimation, because it estimates the root depth based on a full image; while the root depth of the top-down network is estimated based on an individual skeleton. 
Our integration network demonstrates superior performance compared to hard or linear combining the poses from the top-down and bottom-up networks, which validates its effectiveness. 
Moreover, we observe that the proposed semi-supervised learning and test time optimization further improve the performance of 3D multi-person pose estimation. 

\noindent \textbf{Semi-Supervised Learning} Other than validating our top-down and bottom-up networks, we also perform ablation analysis on our semi-supervised learning. In Table~\ref{tab:ssl}, we show the result of using reprojection loss, multi-perspective loss, reprojection loss with the proposed discriminator, and reprojection + multi-perspective loss with discriminator. We find that the reprojection loss is more useful than the multi-perspective loss because it leverages the information from the 2D pose estimator, which is trained with 2D human pose datasets with larger variations in human pose and appearance . More importantly, we observe that our proposed interaction-aware discriminator demonstrates the largest performance improvement compared with the other modules, proving the importance of enforcing the validity of the interaction between persons. 
Note that, since \mbox{Table~\ref{tab:ssl}} is focused on evaluating the semi-supervised learning part, the test time optimization is not used in this experiment.

\noindent \textbf{Inter-Person Discriminator} 
To validate our choice in the inter-person discriminator to check the interaction between two persons, we conduct parameter analysis in terms of the number of poses to check in the discriminator as illustrated in Table~\ref{tab:discrim}. Checking the pose validity for only one person (results of Dis1), which is what most existing methods do, shows improvement against baseline. However, when checking the validity of pair of poses (results of Dis2), we observe the accuracy improves in both person-centric and camara-centric pose estimation metrics. Further improving the number of poses to three does not help but negatively affect the performance. Thus, this analysis supports the use of the pair of poses in validity checking for the inter-person discriminator. 
Note that, since \mbox{Table~\ref{tab:discrim}} is focused on evaluating the inter-person discriminator part, the test time optimization is not used in this experiment.

\noindent \textbf{Test Time Optimization} 
We also provide ablation study results of the test time optimization to understand the contribution of each loss term as shown in Table~\ref{tab:tto}.
The baseline method in the table is the GCN network\mbox{\cite{cheng2021graph}} with 2D ground truth (w. GT) or 2D pose estimation \mbox{\cite{sun2019hrnet}} (w/o GT). The first column in \mbox{Table~\ref{tab:tto}} shows the results of using the ground truth of 2D pose; the second column illustrates the results of using 2D estimator instead. In both cases, we observe that our TTO regularization is helpful to reduce the error on top of the baseline.
In particular, it is observed that the reprojection regularization brings the largest error reduction compared with the baseline. By adding each additional regularization or optimization strategy, the error is further reduced, combining all the losses and strategies achieves the highest performance in terms of MPJPE metric, which validates the proposed test time optimization is helpful to refine the estimated 3D poses at the inference stage. 

\noindent \textbf{Temporal Window Analysis} 
Finally, we analyze the performance of different temporal window lengths for different motion trajectory order in our trajectory regularization on the Human3.6M dataset using ground-truth 3D poses as shown in Table \ref{tab:traj}. It is observed that using temporal windows 2, 5, 5 for order 1, 2, 3 achieves the lowest error, which are the parameters we choose for the following experiments. 
Same as before, since \mbox{Table~\ref{tab:traj}} is focused on evaluating the temporal window lengths, TTO is not used in this experiment.

\noindent \textbf{Runtime Analysis} Table \mbox{\ref{tab:runtime}} shows the runtime of each component of the proposed method. We use HRNet-w32 \mbox{\cite{sun2019hrnet}} as the backbone network for both multi-person pose estimators in the top-down and bottom-up networks. The input resolution for bottom-up network is $512$ and $256$ for multi-person pose estimator in top-down branch. For object detector, we use Faster RCNN \mbox{\cite{ren2015faster}} with ResNeXt \mbox{\cite{Xie2016}} backbone. The temporal window length is set to $243$ for TCN. For the TTO, we run both steps with 3000 iterations on the whole video and compute the runtime per frame. The runtime speed is tested on single RTX2080Ti GPU with I7-9900k CPU. 

\begin{table}[]
    \centering
    \begin{tabular}{c|c}

    \cline{1-2}
    \cline{1-2}
    \rule{0pt}{2.6ex}
         Method & Time(ms)  \\
         \cline{1-2}
         \rule{0pt}{2.6ex}
         Human detector & 97 \\
         2D pose estimator & 53\\
         GCN & 4\\
         TCN & 9\\
         Bottom-up & 201 \\
         Integration & 6\\
         TTO & 31 \\
         \cline{1-2} 
         \rule{0pt}{2.0ex}
         Total & 401\\
    \cline{1-2}
    \cline{1-2}
    
    \end{tabular}
    \vspace{0.5em}
    \caption{Runtime of each component.}
    \label{tab:runtime}
\end{table}

\vspace{-1.0em}
\subsection{Quantitative Evaluation}


\begin{table}[t]
	\footnotesize
	\centering
	\begin{tabular}{c|c|c|c}
		\cline{1-4}
		\cline{1-4}
		\rule{0pt}{2.6ex}
		\textbf{Group} & \textbf{Method} & PCK & PCK$_{abs}$\\
		\cline{1-4}
		\rule{0pt}{2.6ex}
		& Mehta et al. \cite{mehta2018single} & 65.0 & n/a\\
		Person- & Rogez et al., \cite{rogez2019lcr} & 70.6 & n/a\\
		centric & Cheng et al. \cite{cheng2019occlusion} & 74.6 & n/a\\
		& Cheng et al. \cite{cheng2020sptaiotemporal} & 80.5 & n/a\\
		\cline{1-4}
		& Moon et al. \cite{Moon_2019_ICCV_3DMPPE} & 82.5 & 31.8\\
		
		Camera- & Lin et al. \cite{lin2020hdnet} & 83.7 & 35.2\\
		centric & Zhen et al. \cite{zhen2020smap} & 80.5 & 38.7\\
		& Li et al. \cite{li2020hmor} & 82.0 & 43.8\\
		& Cheng et al. \cite{cheng2021graph} & \underline{87.5} & \underline{45.7}\\ 
		& Our method & \textbf{89.6} & \textbf{48.1}\\
		\cline{1-4}
		\cline{1-4}
	\end{tabular}
	\vspace{0.5em}
	\caption{Quantitative evaluation on multi-person 3D dataset, MuPoTS-3D. Best in \textbf{bold}, second best \underline{underlined}. }
	\vspace{-0.5em}
	\label{tab:MuPoTS_3d}
\end{table}

\begin{table}[h]
	\footnotesize
	\centering
	\begin{tabular}{c|c|c|c}
		\cline{1-4}
		\cline{1-4}
		\rule{0pt}{2.6ex}
		\textbf{Method} & $t=0.4m$ & $t=0.8m$ & $t=1.2m$ \\
		\cline{1-4}
		\rule{0pt}{2.6ex}
		\cite{redmon2018yolov3} + \cite{martinez2017simple} + \cite{rogez2019lcr} & 39.14 & 47.38 & 49.03\\
		LoCO \cite{fabbri2020compressed} & \underline{50.82} & \underline{64.76} & \underline{70.44}\\
		Ours & \textbf{58.15} & \textbf{69.32} & \textbf{74.19}\\
		\cline{1-4}
	\end{tabular}
	\vspace{0.5em}
	\caption{Quantitative results on JTA dataset. F1 values are reported based on different threshold $t$ when the point is considered "true positive" when the distance from corresponding distance is less than $t$. Best in \textbf{bold}, second best \underline{underlined}.}
    \vspace{-0.5em}
	\label{tab:jta}
\end{table}

To evaluate the performance for 3D multi-person camera-centric pose estimation in both indoor and outdoor scenarios, we perform evaluations on MuPoTS-3D as summarized in Table~\ref{tab:MuPoTS_3d}. The results show that our camera-centric multi-person 3D pose estimation outperforms the SOTA \cite{li2020hmor} on $PCK_{abs}$ by $2.3\%$. We also perform person-centric 3D pose estimation evaluation using $PCK$ where we outperform the SOTA method \cite{lin2020hdnet} by $2.1\%$. The evaluation on MuPoTS-3D shows that our method outperforms the state-of-the-art methods 
in both camera-centric and person-centric 3D multi-person pose estimation as our framework overcomes the weaknesses of both bottom-up and top-down networks and at the same time benefits from their strengths. 
Note that, our method in \mbox{Table~\ref{tab:MuPoTS_3d}} means our full model, including the test time optimization, same for the following evaluations in Table \mbox{\ref{tab:jta}, \ref{tab:h36m}, \ref{tab:3dpw}}.

Following recent work \cite{fabbri2020compressed}, we also perform evaluations on JTA, which is a synthetic dataset acquired from computer games, to further validate the effectiveness of our method for camera-centric 3D multi-person pose estimation. As shown in Table~\ref{tab:jta}, our method is superior over the SOTA method \cite{fabbri2020compressed} (e.g., our result shows $12.6\%$ improvement on F1 value, $t=0.4m$) on this challenging dataset where both inter-person occlusion and large person scale variation present, which again illustrate that our proposed method can handle these challenges in 3D multi-person pose estimation. 

Human3.6M is widely used for evaluating 3D single-person pose estimation. As our method is focused on dealing with inter-person occlusion and scale variation, we do not expect our method to perform significantly better than the SOTA methods. Table~\ref{tab:h36m} summarizes the quantitative evaluation on Human3.6M where our method is comparable with the SOTA methods \cite{kolotouros2019learning,li2020hmor} on person-centric 3D human pose evaluation metrics (i.e., MPJPE and PA-MPJPE). 

3DPW is an outdoor multi-person 3D human shape reconstruction dataset. It is unfair to compare the errors between skeleton-based methods with ground-truth defined on SMPL model~\cite{loper2015smpl} due to the different definitions of joints~\cite{tripathi2020posenet3d}. We run human detection on all frames and create an occlusion subset where the frames with the large overlay between persons are selected. The performance drop between the full testing test of 3DPW and the occlusion subset can effectively tell if a method can handle inter-person occlusion, which is shown in Table~\ref{tab:3dpw}. We observe that our method shows the least performance drop from the testing set to the subset, which demonstrates our method is indeed more robust to inter-person occlusion.

\begin{table}[t]
	\footnotesize
	\centering
	\begin{tabular}{c|c|c|c}
		\cline{1-4}
		\cline{1-4}
		\rule{0pt}{2.6ex}
		\textbf{Group} & \textbf{Method} & MPJPE & PA-MPJPE\\
		\cline{1-4}
		\rule{0pt}{2.6ex}
		& Hossain et al., \cite{hossain2018exploiting} & 51.9 & 42.0 \\
		& Wandt et al., \cite{wandt2019repnet}* & 50.9 & 38.2 \\
		\small{Person-} & Pavllo et al., \cite{pavllo20193d} & 46.8 & 36.5 \\
		\small{centric} & Cheng et al., \cite{cheng2019occlusion} & 42.9 & 32.8\\
		& Kocabas et al., \cite{kocabas2020vibe} & 65.6 & 41.4  \\
		& Kolotouros et al. \cite{kolotouros2019learning} & n/a & \underline{41.1} \\
		\cline{1-4}
		& Moon et al., \cite{Moon_2019_ICCV_3DMPPE} & 54.4 & 35.2 \\
		\small{Camera-} & Zhen et al., \cite{zhen2020smap} & 54.1 & n/a \\
		\small{centric} & Li et al., \cite{li2020hmor} & 48.6 & \underline{30.5} \\
		& Ours & \textbf{39.1} & \textbf{29.3} \\
		\cline{1-4}
	\end{tabular}
	\vspace{0.5em}
	\caption{Quantitative evaluation on Human3.6M for normalized and camera-centric 3D human pose estimation. * denotes ground-truth 2D labels are used. Best in \textbf{bold}, second best \underline{underlined}. }
	\label{tab:h36m}
	\vspace{-0.5em}
\end{table}

\begin{table}
	\footnotesize
	\centering
	\begin{tabular}{c|c|c|c}
		\cline{1-4}
		\cline{1-4}
		\rule{0pt}{2.6ex}
		\textbf{Dataset} & \textbf{Method} & PA-MPJPE & $\delta$ \\
		\cline{1-4}
		\rule{0pt}{2.6ex}
		& Doersch et al. \cite{doersch2019sim2real}  & 74.7 & n/a \\
		& Kanazawa et al. \cite{humanMotionKanazawa19} & 72.6 & n/a \\
		Original & Arnab et al. \cite{arnab2019exploiting} & 72.2 & n/a \\
		& Cheng et al. \cite{cheng2020sptaiotemporal} & 71.8 & n/a \\
		& Sun et al. \cite{sun2019human} & 69.5 & n/a \\
		& Kolotouros et al. \cite{kolotouros2019learning}* & \underline{59.2} & n/a \\
		& Kocabas et al., \cite{kocabas2020vibe}* & \textbf{51.9} & n/a \\
		& Our method  & 61.7 & n/a \\
		\cline{1-4}
		& Cheng et al. \cite{cheng2020sptaiotemporal} & 92.3 & +20.5\\
		& Sun et al. \cite{sun2019human} & 84.4 & \underline{+14.9}\\
		Subset & Kolotouros et al. \cite{kolotouros2019learning}*  & 79.1 & +19.9\\
		& Kocabas et al., \cite{kocabas2020vibe}* & \textbf{72.2} & +20.3 \\
		& Our method  & \underline{72.4} & \textbf{+9.5}\\
		\cline{1-4}
	\end{tabular}
	\vspace{0.5em}
	\caption{Quantitative evaluation using PA-MPJPE on original 3DPW test set and its occlusion subset. * denotes extra 3D datasets were used in training. Best in \textbf{bold}, second best \underline{underlined}.}
	\vspace{-1em}
	\label{tab:3dpw}
\end{table}

\begin{figure*}[t]
	\centering
	\makebox[\textwidth]{\includegraphics[width=\textwidth]{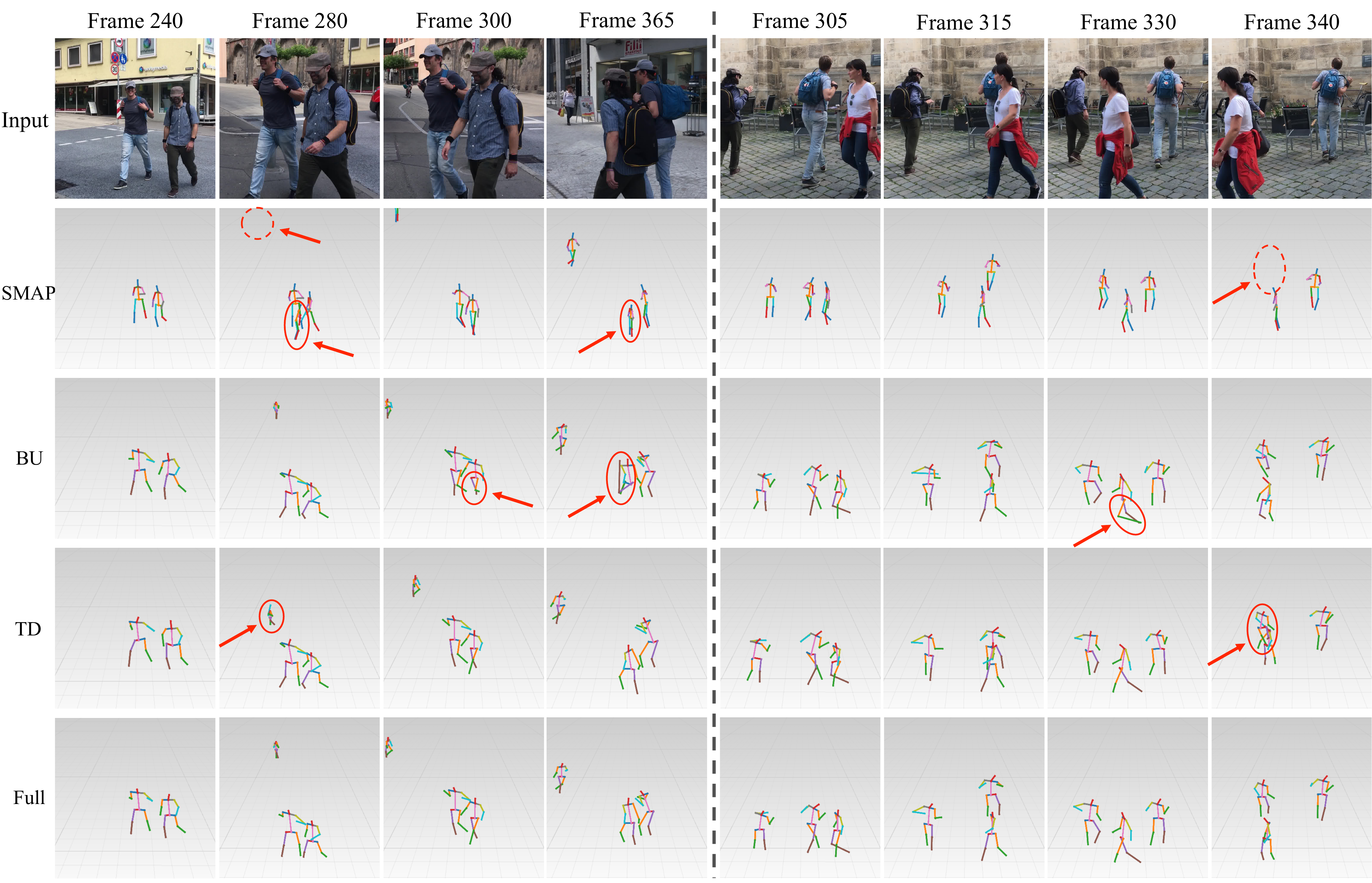}}
	\caption{Examples of results from our whole framework compared with different baseline results on two video clips on 3DPW dataset. First row shows the images from two video clips; second row shows the results from SMAP \cite{zhen2020smap}; third row shows the result of of our bottom-up (BU) network; fourth row shows the results of our top-down (TD) network; last row shows the results of our full model. Wrong estimations are labeled with red circles.}
	\label{fig:qualitative_evaluation}
	\vspace{-1em}
\end{figure*}

\vspace{-1.0em}
\subsection{Qualitative Evaluation}

Fig. \ref{fig:qualitative_evaluation} shows the comparison among a SOTA bottom-up method SMAP \cite{zhen2020smap}, our bottom-up network, top-down network, and full model. 
We observe that SMAP suffers from person scale variation where the person who is far from the camera is missing in frame 280 as well as inter-occlusion (e.g., frame 365 and 340). 
Our bottom-up network is robust to scale variance, but fragile to the out-of-image poses as our discriminator is not used here (e.g., frame 365 and 330). 
Moreover, our top-down network produces reasonable relative poses with the aid of GCN and TCNs. 
However, there exists an error of camera-centric root depth in our top-down network, because our top-down network estimates root depth based on individual 2D poses and lacks global awareness (e.g., frame 280). 
Finally, our full model benefits from both networks and produces the best 3D pose estimations among these baselines. 

\begin{figure*}
    \centering
    \includegraphics[width=\textwidth]{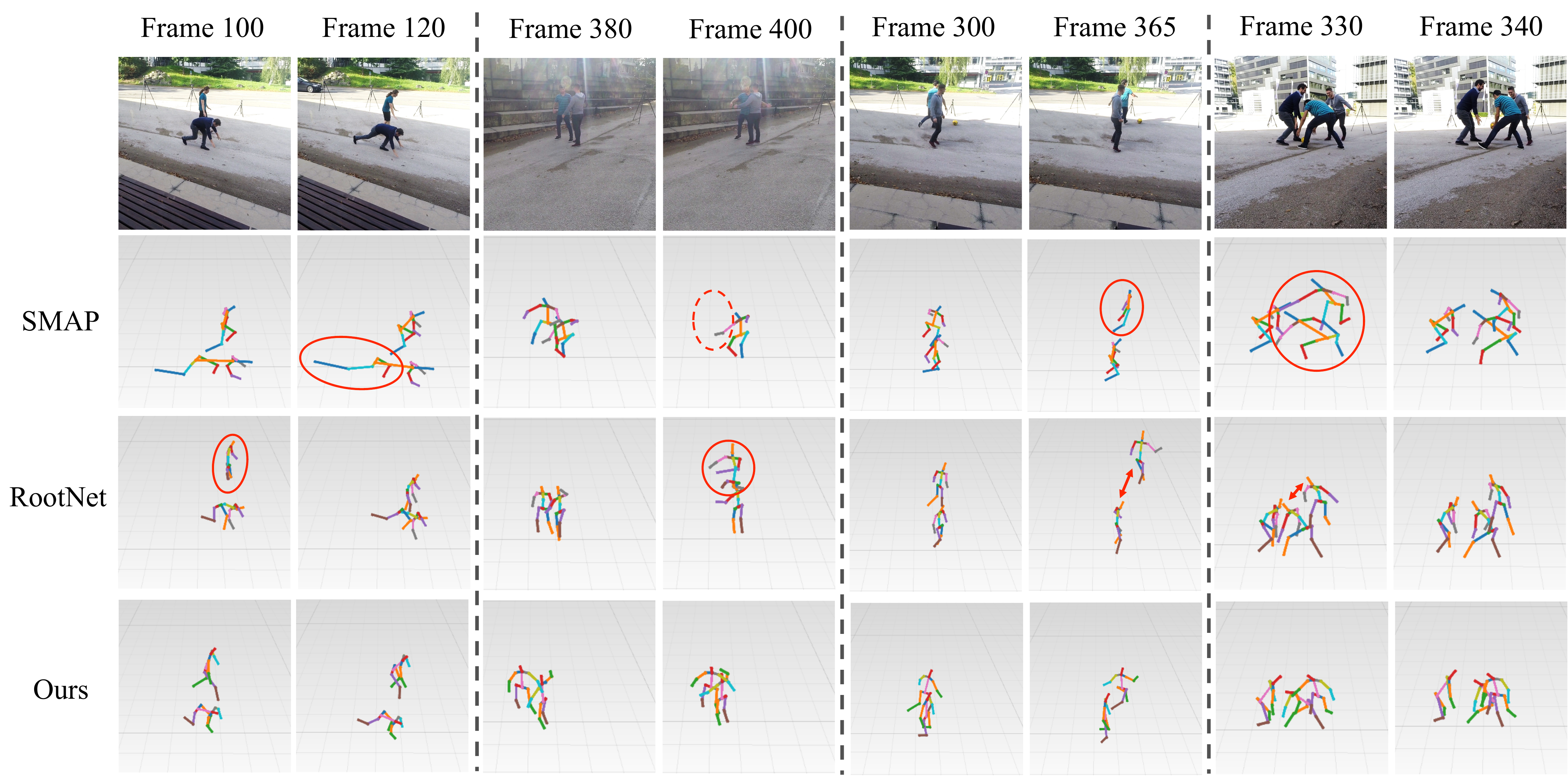}
    \caption{Results of our method compared with that of SMAP \cite{zhen2020smap} (i.e., the SOTA bottom-up method) and RootNet \cite{Moon_2019_ICCV_3DMPPE} (i.e., the SOTA top-down method) on MuPoTS dataset. Results from four video clips are included: top-left, top-right, bottom-left, and bottom-right. For each video clip, the first row is the frames from the video clip; the second row is the result of SMAP; the third row is the result of RootNet; the fourth row is the result of our method. It is observed from these results that the SOTA methods suffer from inter-person occlusions while our method can handle these challenges and produce accurate camera-centric 3D multi-person pose estimation.}
    \label{fig:mupots}
\end{figure*}

\begin{figure*}
    \centering
    \includegraphics[width=0.9\textwidth]{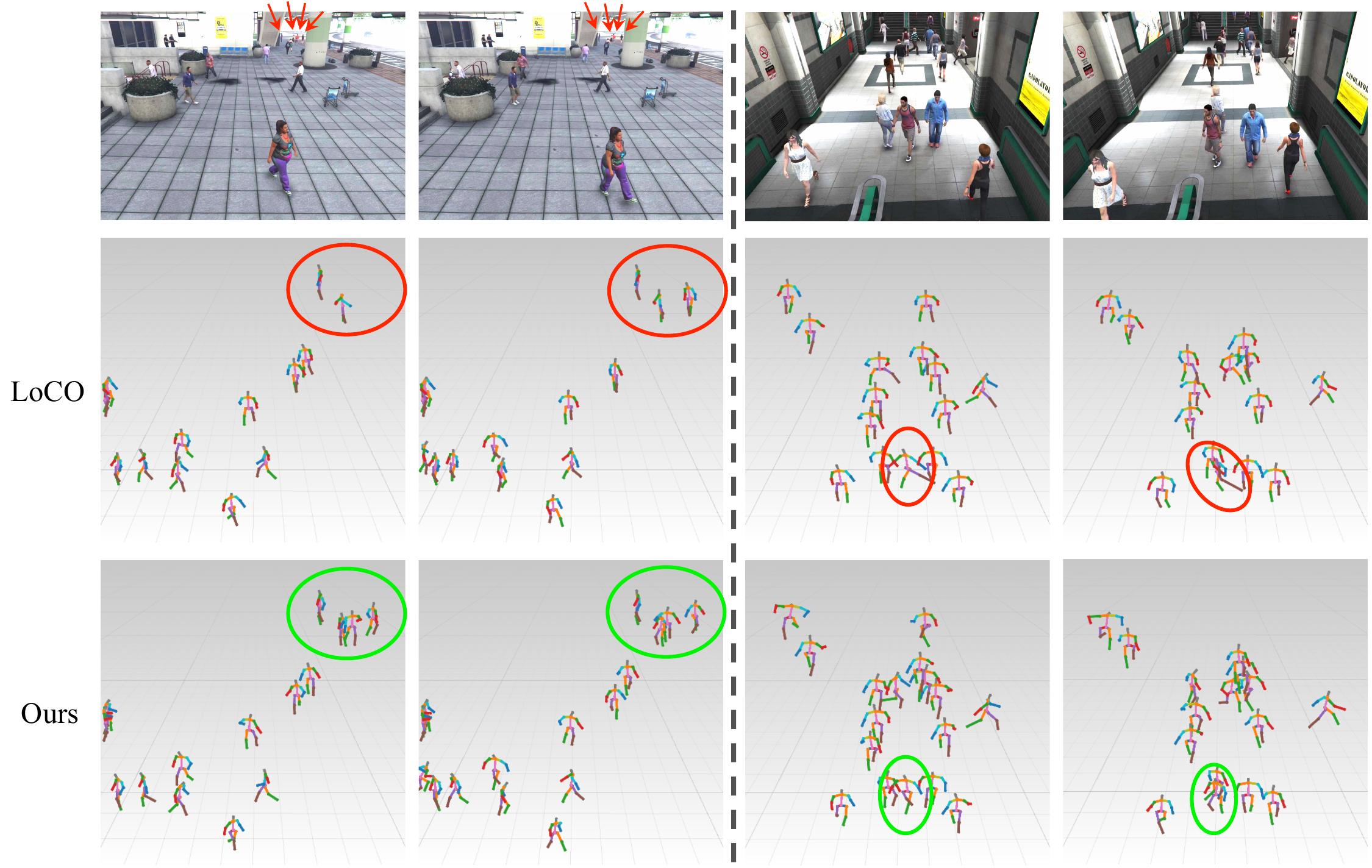}
    \caption{Result of our method compared with that of LoCO \cite{fabbri2020compressed} (i.e., a SOTA method released trained model on JTA) on JTA dataset. Results from two video clips are included: top and bottom separated by the dashed line. For each video clip, the first row is the frames from the video clip; the second row is the result of LoCO; the third row is the result of our method. These results show that on this synthetic datasets, our method is able to produce more accurate and robust 3D multi-person pose estimation compared with other methods. We use red circle to indicate the wrong results of LoCO and green circle to point out the corresponding correct results of our method. In the first row of the top video clip, due the four persons are far from the camera which are small, we use four red arrows to indicate each of them.}
    \label{fig:jta}
\end{figure*}

To further demonstrate the performance of our method compared with the SOTA 3D multi-person pose estimation methods, we provide additional qualitative results of our method compared with that of the SOTA bottom-up method SMAP \cite{zhen2020smap} and the SOTA top-down method RootNet \cite{Moon_2019_ICCV_3DMPPE} on four video clips from MuPoTS dataset, as shown in Fig. \ref{fig:mupots}. The errors of existing methods are highlighted in red circles or arrows, where we clearly observe our method outperforms the existing methods and can provide accurate camera-centric 3D multi-person pose estimation when inter-person occlusions present in the videos. Moreover, as we reported our quantitative performance on JTA dataset in Table \ref{tab:jta}, we also provide the qualitative results of our method compared with that of the SOTA method reported and released their trained model on the JTA dataset \cite{fabbri2020compressed} in Fig. \ref{fig:jta}. The two video clips in Fig. \ref{fig:jta} show both inter-person occlusions and large multi-person scale variation where we observe our method can handle both challenges well and produce accurate camera-centric 3D multi-person pose estimation compared with LoCO \cite{fabbri2020compressed}. 

\begin{figure}
    \centering
    \includegraphics[width=\linewidth]{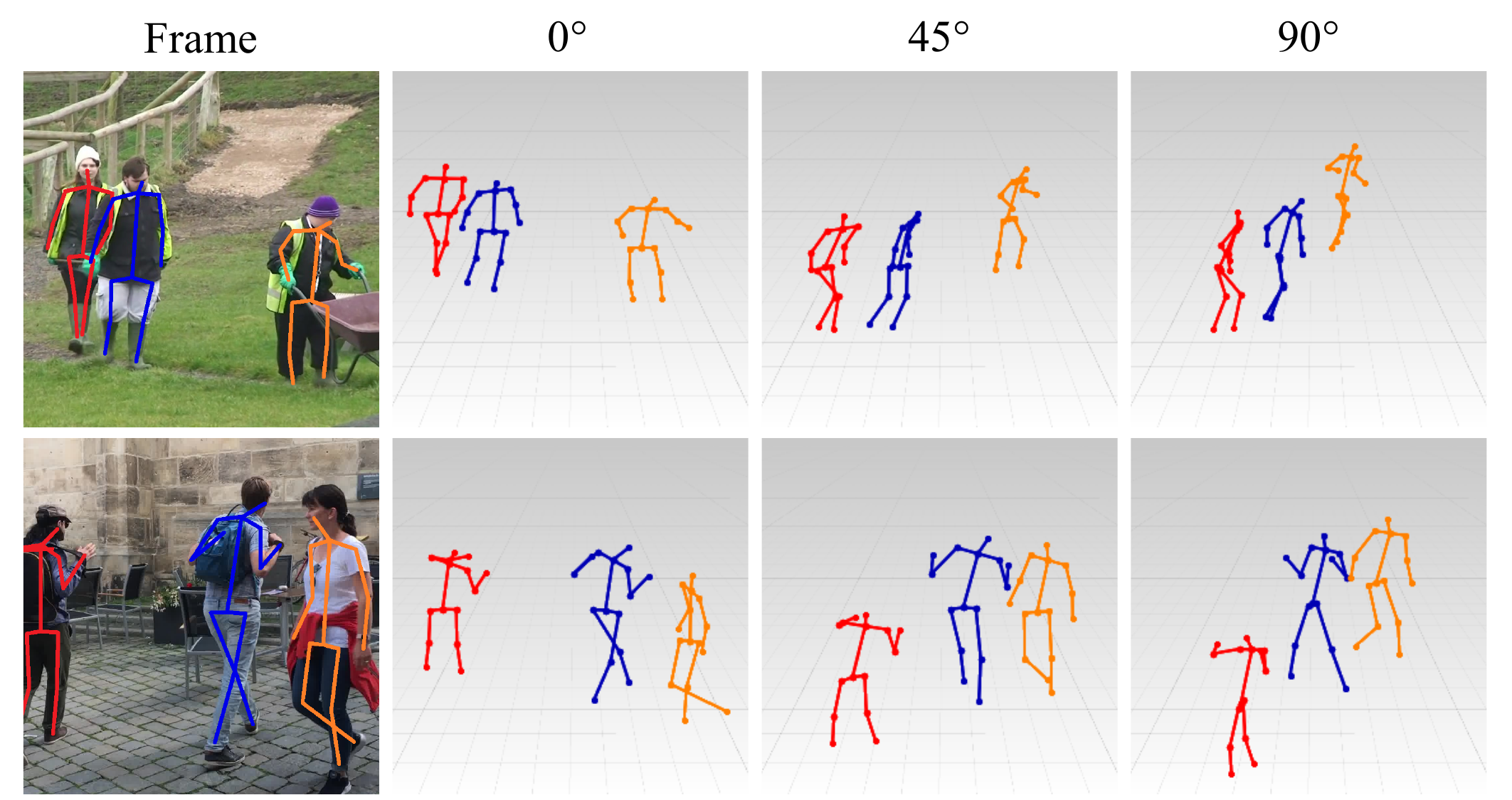}
    \caption{Qualitative results of the estimated 2D poses overlaying on input images and the estimated 3D poses visualized in novel viewpoints (virtual camera rotated by 0, 45, 90 degrees clockwise).
    Different colors are used for different persons in both 2D and 3D human poses for better visualization purpose. Top frame from Posetrack dataset, bottom example from 3DPW dataset.}
    \label{fig:multi_persp}
    \vspace{-0.8em}
\end{figure}

\begin{figure}[t]
    \centering
    \includegraphics[width=\linewidth]{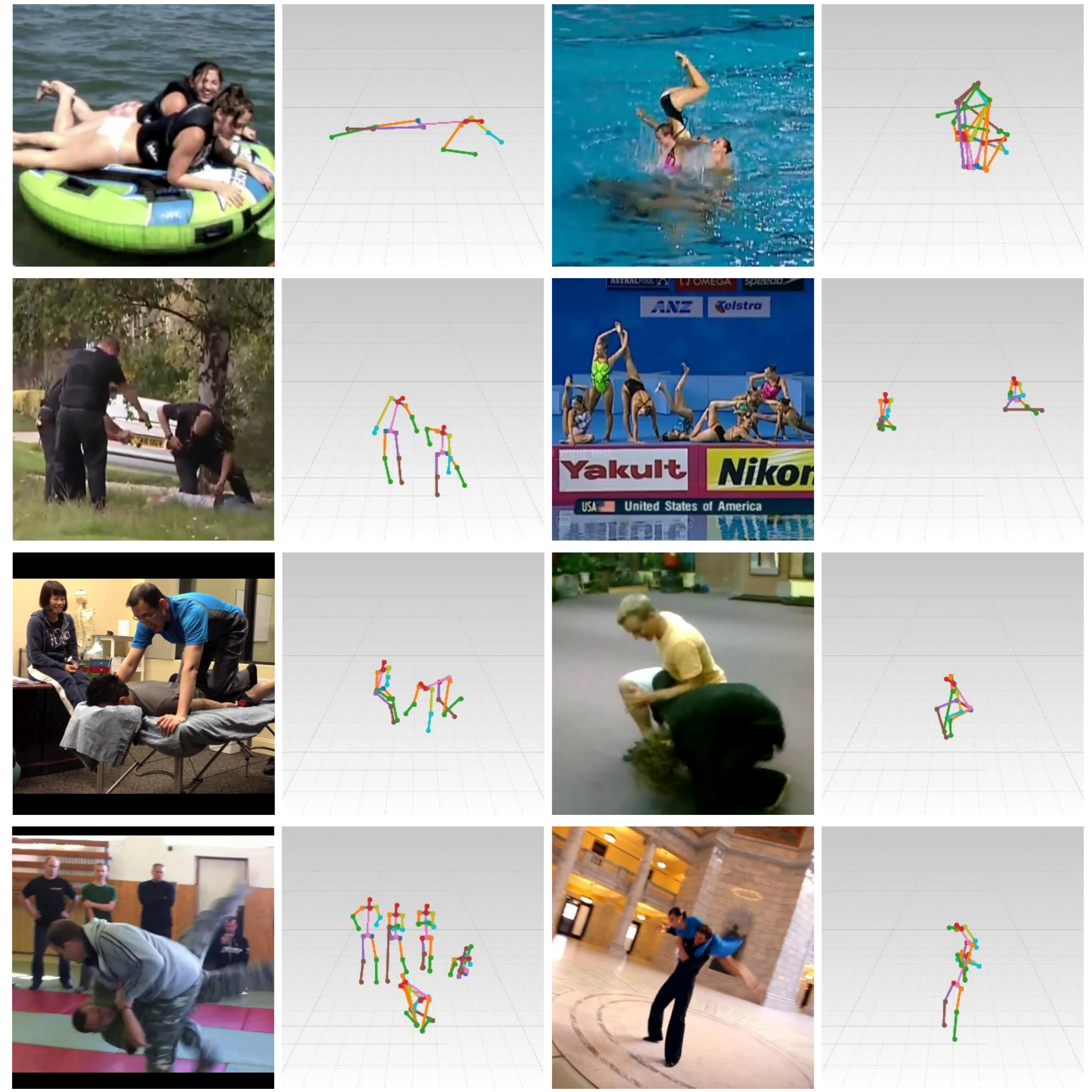}
    \caption{Failure cases of our method. Example frames are from PoseTrack dataset.
    }
    \label{fig:failure}
    \vspace{-0.8em}
\end{figure}

We also provide results of the estimated 3D poses in novel viewpoints and the estimated 2D poses overlaid on input images as in Fig.~\ref{fig:multi_persp}. Our estimated camera-centric 3D poses visualized from different angles further validate the effectiveness of the proposed method, where the depth of each person is estimated reasonably well from the results of different angles. 

\begin{figure*}
    \centering
    \includegraphics[width=\textwidth]{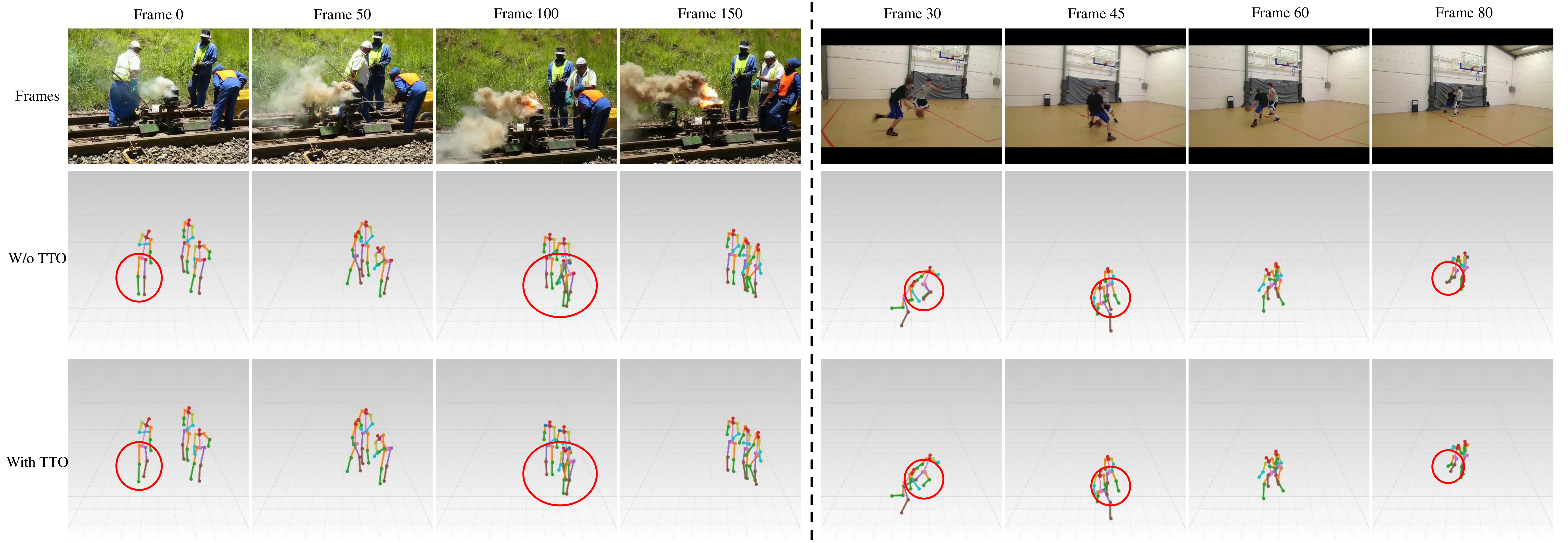}
    \caption{Qualitative results of TTO module. Major improvements of the TTO module are highlighted in red circles. Video clips are from PoseTrack dataset.}
    \label{fig:tto}
\end{figure*}

Qualitative comparisons of the proposed TTO module are shown in \mbox{Fig.~\ref{fig:tto}}, where results of our full model with and without TTO on two wild videos are provided. In both video clips, we observe that the predicted 3D human poses without TTO are inaccurate (i.e., highlighted in red circles), which are affected by poor illumination or occlusion. In contrast, the results with TTO are improved, where the inaccuracies are fixed. Thus, these qualitative results demonstrate the effectiveness of the proposed TTO module on wild video.

\subsection{Failure Analysis}

Fig. \mbox{\ref{fig:failure}} shows the failure cases where the images are from the PoseTrack dataset \mbox{\cite{iqbal2017posetrack}}. There are two major kinds of failure cases of our method on the wild videos. First, the persons that are constantly heavily occluded in a video are likely to be missed by our method, as shown in the left column of Fig. \mbox{\ref{fig:failure}}.
In such cases, the occluded person usually has only few joints with high confidence across frames, which makes it difficult to group them together to form a complete human pose. Thus, robustly dealing with incomplete or even extremely incomplete human poses is still a challenge to be solved in the future.
A second kind of failure cases is the extreme human poses, as shown in the right column of Fig. \mbox{\ref{fig:failure}}. The extreme human poses are rare, which are usually not well represented in the training dataset. This can be viewed as an out-of-distribution problem, where the extreme human poses are out of the distribution of the training data (i.e., training dataset is dominated by regular or normal human poses.). As a result, both 2D and 3D pose estimation cannot handle these images well. Addressing this issue is another future work.

\section{Conclusion}
We have proposed a novel method for monocular-video 3D multi-person pose estimation, which addresses the problems of inter-person occlusion and close interactions.
We introduced the integration of top-down and bottom-up approaches to exploit their strengths. 
A novel inter-person pose discriminator is proposed to enforce the validity of human poses of close pairwise interactions. 
Semi-supervised learning and test time optimization are proposed to further improve the accuracy of 3D multi-person pose estimation. 
In addition, we introduce unsupervised losses, i.e. the high-order temporal constraint, reprojection loss, and bone-length regularization.,which enable the optimization in test time. This optimization is critical in improving our performance, particularly in the cases where the gaps between the training and testing data are relatively significant.
Our quantitative and qualitative evaluations show the effectiveness of our method compared to the state-of-the-art baselines.

%



\ifCLASSOPTIONcompsoc
  \section*{Acknowledgments}
\else
  \section*{Acknowledgment}
\fi

This research is supported by the National Research Foundation, Singapore under its Strategic Capability Research Centres Funding Initiative. Any opinions, findings and conclusions or recommendations expressed in this material are those of the author(s) and do not reflect the views of National Research Foundation, Singapore.

\bibliographystyle{IEEEtran}
\bibliography{egbib}

\begin{IEEEbiography}[{\includegraphics[width=1in,height=1.25in,clip,keepaspectratio]{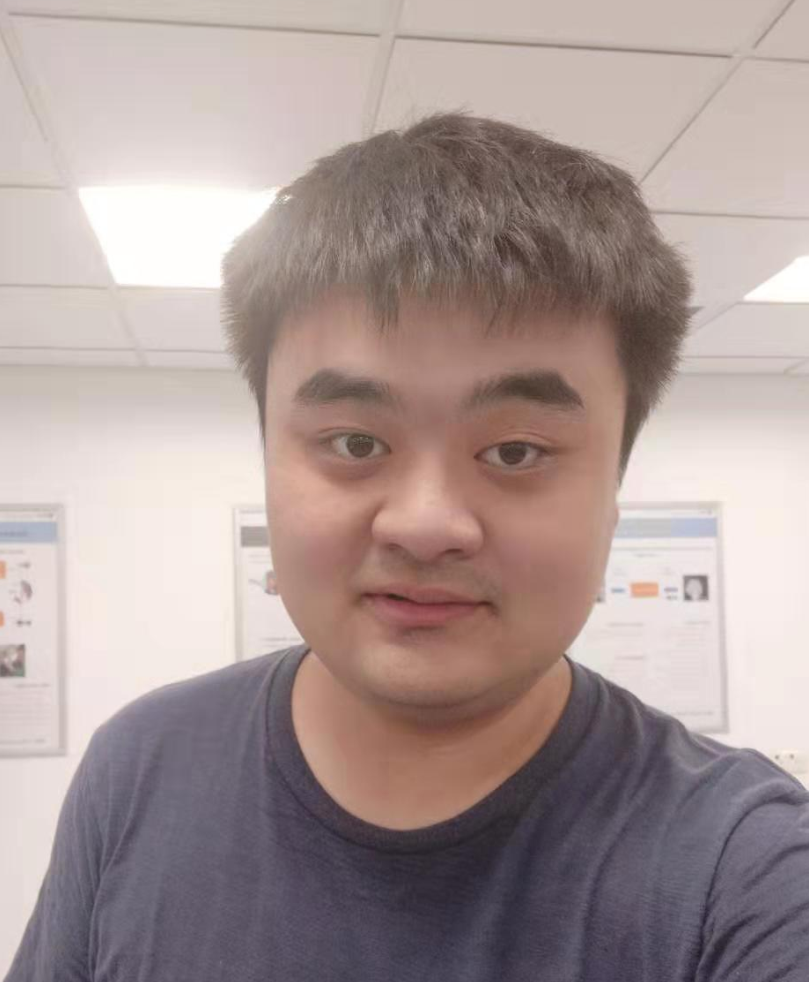}}]{Yu Cheng}
received the bachelor's degree in Electrical and Electronic Engineering from Nanyang Technological University, Singapore, in 2018. He is now a Ph.D. candidate in ECE (Electrical and Computing Engineering) in National University of Singapore. His research interests include human pose estimation, facial recognition and object detection. 
\end{IEEEbiography}

\begin{IEEEbiography}[{\includegraphics[width=1in,height=1.25in,clip,keepaspectratio]{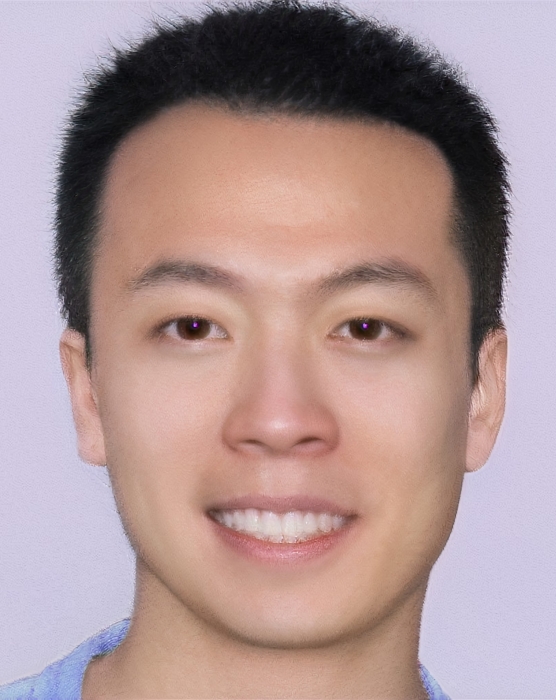}}]{Bo Wang} received the PhD degree from School of Computing at the University of Utah in 2015. He is currently with CtrsVision. He was previously with GE Global Research and Tencent America. His research interests include 3D computer vision and machine learning with application to human pose estimation, spatial and temporal visual data modeling and analysis.
\end{IEEEbiography}

\begin{IEEEbiography}[{\includegraphics[width=1in,height=1.25in,clip,keepaspectratio]{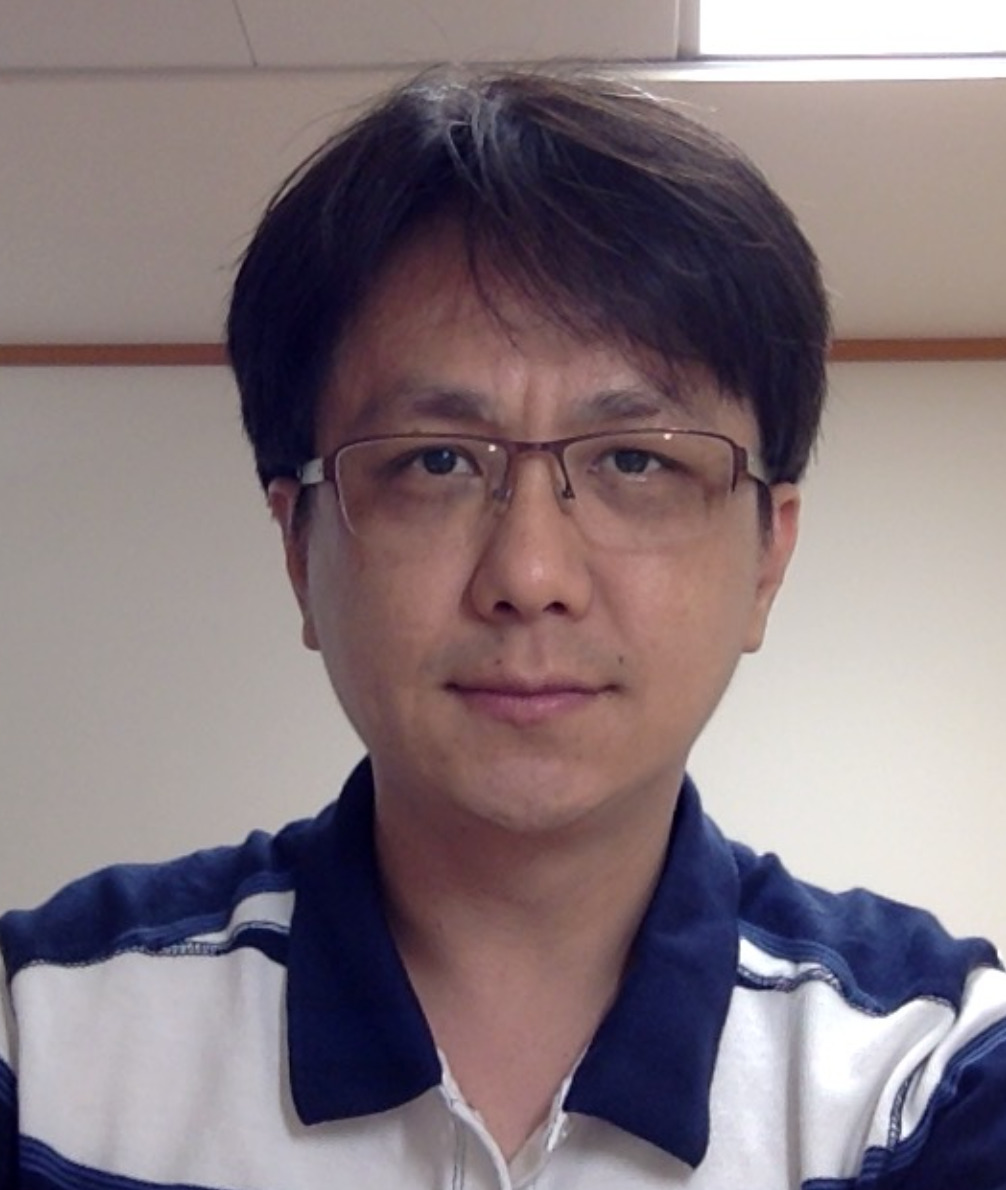}}]{Robby T. Tan}
received the PhD degree in computer science from the University of Tokyo. He is now an associate professor at both Yale-NUS College and ECE (Electrical and Computing Engineering), National University of Singapore. His research interests include computer vision and deep learning, particularly in the domains of low level vision (bad weather/nighttime, color analysis, physics-based vision, optical flow, etc.), human pose/motion analysis, and applications of deep learning in healthcare. He is a member of the IEEE.
\end{IEEEbiography}

\vfill







\end{document}